\documentclass{article}

\usepackage[utf8]{inputenc}
\usepackage[T1]{fontenc}
\usepackage{textcomp}
\usepackage{times}
\usepackage{iclr_conference}
\iclrfinalcopy
\usepackage{booktabs}
\usepackage{nicefrac}
\usepackage{microtype}
\usepackage{graphicx}
\usepackage{float}
\usepackage{amsmath,amsfonts,bm}

\def\eqref#1{equation~\ref{#1}}

\def\1{\bm{1}}

\def\vb{{\bm{b}}}

\def\vd{{\bm{d}}}

\def\vw{{\bm{w}}}
\def\vx{{\bm{x}}}
\def\vy{{\bm{y}}}
\def\vz{{\bm{z}}}

\def\mW{{\bm{W}}}

\DeclareMathAlphabet{\mathsfit}{\encodingdefault}{\sfdefault}{m}{sl}
\SetMathAlphabet{\mathsfit}{bold}{\encodingdefault}{\sfdefault}{bx}{n}

\def\sF{{\mathbb{F}}}

\newcommand{\Wenc}{\mW_{\text{enc}}}
\newcommand{\Wdec}{\mW_{\text{dec}}}

\newcommand{\bpre}{\vb_{\text{pre}}}
\newcommand{\benc}{\vb_{\text{enc}}}
\newcommand{\xhat}{\hat{\vx}}

\usepackage{amssymb}
\usepackage{xcolor}
\usepackage[font=small,labelfont=bf,labelsep=period]{caption}
\usepackage{appendix}
\usepackage{titletoc}
\usepackage{tcolorbox}
\tcbuselibrary{skins,breakable,listings}
\newtcolorbox{importantbox}[1][Important Note]{
    enhanced,
    colback=blue!5,
    colframe=blue!75!black,
    arc=4mm,
    boxrule=1.5pt,
    title=#1,
    fonttitle=\bfseries,
    attach boxed title to top left={yshift=-3mm, xshift=5mm},
    coltitle=black,
    boxed title style={
        enhanced,
        colback=blue!40,
        arc=1mm
    },
    top=15pt,
    bottom=10pt
}

\newtcblisting{promptlisting}{
  listing only,
  breakable,
  enhanced,
  colback=black!3,
  colframe=black!35,
  boxrule=0.35pt,
  arc=0.8mm,
  left=1mm,
  right=1mm,
  top=1.5mm,
  bottom=1.5mm,
  before skip=\smallskipamount,
  after skip=\smallskipamount,
  listing options={
    basicstyle=\ttfamily\small,
    breaklines=true,
    breakatwhitespace=false,
    breakindent=0pt,
    xleftmargin=0pt,
    framexleftmargin=0pt,
    columns=fullflexible,
    keepspaces=true,
    aboveskip=0.25\smallskipamount,
    belowskip=0.25\smallskipamount,
  },
}

\definecolor{linkblue}{rgb}{0,0.08,0.45}
\usepackage[
  colorlinks=true,
  linkcolor=linkblue,
  citecolor=linkblue,
  urlcolor=linkblue,
  pdfborder={0 0 0},
  pdftitle={RouterInterp: Understanding Superposed Specialisation in Mixture of Experts Routing},
  pdfauthor={Ilya Lasy, Nora Yinuo Cai, Kola Ayonrinde},
  pdfsubject={Published at the 43rd International Conference on Machine Learning (ICML 2026)}
]{hyperref}
\usepackage{xurl}
\usepackage{cleveref}

\title{RouterInterp: Understanding Superposed\\Specialisation in Mixture of Experts Routing}

\author{%
  Ilya Lasy\\
  Faculty of Informatics, TU Wien\\
  \texttt{ilya.lasy@tuwien.ac.at}\\
  \And
  Nora Yinuo Cai\\
  Independent\\
  \And
  Kola Ayonrinde\\
  UK AI Security Institute\\
  \texttt{koayon@gmail.com}\\
}

\date{}

\begin{document}

\maketitle

\begin{abstract}
  Sparse Mixture of Experts (MoE) models scale more efficiently than dense models
  by routing tokens to modular expert networks that are only active
  for processing a fraction of tokens.
  A leading hypothesis for the performance of MoE models is that each expert
  specialises in a single, coherent domain.
  However, interpretability efforts that assume this hypothesis have generally been unsuccessful.
  We propose and present evidence for an alternative account that we call the
  \emph{Superposed Specialisation Hypothesis} (SSH):
  experts specialise in a disjoint union of fine-grained features rather than one broad domain.
  Leveraging the SSH, we introduce \emph{RouterInterp},
  a method for interpreting expert routing that identifies Sparse Autoencoder features
  most predictive of routing decisions and produces unified natural language explanations.
  On gpt-oss-20b, RouterInterp explains expert routing with ${\sim}65\%$
  higher detection accuracy than prior token statistics based methods.
  This work provides a scalable method for generating more accurate explanations
  of expert routing and increases our understanding of a previously uninterpretable
  component of foundation models.\footnote{
      Code available at \url{https://github.com/ilyalasy/routerinterp}.
  }

\end{abstract}

\section{Introduction}
\label{sec:intro}
Sparse Mixture-of-Experts (MoE) transformers have emerged as a promising approach for scaling
frontier language models \citep{cai2025survey, fedus2022switch, du2022glam}.
In dense (i.e. non-MoE) transformers \citep{radford2019gpt2}, each input is processed by all the parameters.
In contrast, MoE models contain multiple expert networks
and a routing mechanism that selects a subset of these ``experts''
for each input token at each layer.
\citet{shazeer2017sparse_moe} and \citet{liu2024deepseek_v3}
show that sparse MoE models can activate only 2–15\% of parameters per input,
enabling dramatic parameter scaling with minimal increases
in inference cost and latency, leading to strong performance on a variety of tasks
\citep{fedus2022moe_review}.

The strong performance of Sparse MoE models has often been
attributed to expert specialisation \citep{lewis2021base_moe}:
if each expert learns to handle a subset of the input data distribution
or perform only a subset of computations,
then using only a subset of the parameters for each input can be both \emph{effective} and \emph{efficient}\footnote{
  \emph{Effective} because all the processing a token requires
  is handled by the selected experts, and \emph{efficient} as only a small fraction of the total parameters and
  associated FLOPs are used for each forward pass.
}.
We call this explanation for the success of MoEs
the \textbf{Specialisation Hypothesis}.

\begin{importantbox}[The Specialisation Hypothesis]
  Sparse Mixture-of-Experts models work well because
  different experts specialise in different parts of the input data distribution
  or different types of computation,
  allowing each input to be processed by only a small subset of the model's parameters.

\end{importantbox}

One seemingly natural corollary of the Specialisation Hypothesis is that
if each expert specialises in a particular domain
then routing decisions should be human-interpretable.
For example, perhaps one expert specialises in processing medical text
and contributes medical knowledge and reasoning when the input is related to healthcare.
Another expert might specialise in generating mathematics and yet another in
storytelling or translating romance languages \citep{ayonrinde2023moe_analogy}.
However, prior work analysing MoE models
has struggled to recover clear, interpretable patterns of
expert specialisation \citep{jiang2024mixtral,lewis2021base_moe,zoph2022st_moe}.
We believe that this difficulty derives from not distinguishing between two
distinct forms of the Specialisation Hypothesis which we term
the \textbf{Domain Specialisation Hypothesis} (DSH) and the
\textbf{Superposed Specialisation Hypothesis} (SSH), respectively.
The distinction between these hypotheses becomes apparent
when we consider that
real-world data contains far more fine-grained categories
(e.g., specific topics, syntactic constructions, reasoning patterns)
than any MoE layer has experts.\footnote{
  Typical MoE models use between 8 and 128 experts per layer \citep{jiang2024mixtral, lewis2021base_moe,zoph2022st_moe}.
}
This means multiple such categories,
which we call \emph{micro-domains}
\footnote{
  For example, ``medical text'' may be a broad domain composed of
  micro-domains such as ``anatomy'', ``diseases'', and ``medical procedures''.
},
must inevitably be routed to the same expert by the Pigeonhole Principle~\citep{rebman1979pigeonhole_principle}.
Under the \emph{Domain Specialisation Hypothesis},
semantically similar micro-domains cluster within an expert,
yielding a coherent domain that the expert specialises in.
We may think of a domain as expressible as a cluster
of similar micro-domains.
In contrast, under the \emph{Superposed Specialisation Hypothesis},
an expert specialises in a disjoint collection of
features spanning multiple unrelated micro-domains.
This mirrors the \emph{superposition} phenomenon observed in neural networks,
where multiple features are represented in the same set of neurons
\citep{elhage2022superposition}.

\begin{importantbox}[Alternative Specialisation Hypotheses]
  \textbf{Domain Specialisation Hypothesis:}
  Sparse Mixture-of-Experts models work well because
  different experts specialise in different \emph{semantically coherent domains}
  (e.g.\ medical text, mathematical reasoning, romance language translation).\\

  \textbf{Superposed Specialisation Hypothesis:}
  Sparse Mixture-of-Experts models work well because
  different experts specialise in \emph{disjoint collections of features}
  corresponding to multiple micro-domains that are \emph{not} highly semantically related.

\end{importantbox}

To test these hypotheses, we need a way to identify the fine-grained features
that drive each expert's activation.
We use sparse autoencoders (SAEs)
\citep{makhzani2013k_sparse_autoencoders, cunningham2024sparse, bricken2023monosemanticity}
to decompose model activations into sparse latent representations (SAE latents),
where each latent corresponds to an interpretable feature.
These SAE latents serve as a feature-level proxy for the micro-domains in our hypotheses.
We find that SAE latents predict which experts are selected far more accurately than
token co-occurrence statistics or sparse linear probes,
achieving a macro-F1 score of 0.73 in routing prediction
for gpt-oss-20b (\Cref{sec:sae_routing_interface}).

Our experiments show that expert routing
can only be explained in terms of multiple semantically disjoint micro-domains,
rather than in terms of a single monosemantic domain (\Cref{sec:spec-hyp:evidence}),
which provides evidence for the SSH.
We suggest that the reason for MoE models preferring superposed specialisation
can be understood through superposition theory
\citep{elhage2022superposition, adler2024complexity_of_comp_in_sup,
hanni2024comp_in_sup, marshall2024polysemanticity_coding_theory}.
Semantically distinct micro-domains co-activate less frequently and hence
there is less interference noise when unrelated domains are co-located
within the same expert, compared to if an expert specialised in a single domain.

Based on these findings, we develop \textbf{RouterInterp},
which interprets routing as arising from a \emph{combination} of SAE features:
it selects the latents most predictive of each expert's routing,
collects inputs where these features and the expert are jointly active,
and explains the expert in natural language
(\Cref{fig:hook_figure}).
To evaluate whether these explanations are human-understandable,
we measure how accurately a language model can predict expert activation
using the explanation (\Cref{sec:methods:scoring}).
On gpt-oss-20b, RouterInterp attains a 0.49 explanation score,
outperforming explanations based on token-expert co-occurrence (0.30).
Our method demonstrates that we can meaningfully understand MoE routing decisions
even when experts specialise in disjoint collections of features.

Our contributions are as follows:

\begin{itemize}
  \item
    We introduce the \emph{Superposed Specialisation Hypothesis (SSH)}: the idea
    that MoE experts specialise in disjoint, semantically heterogeneous collections
    of features rather than a single coherent domain (\Cref{sec:spec-hyp:alternative}).
    We also give two theoretical arguments
    for why this holds in practice (\Cref{sec:spec-hyp:motivation}).
  \item
    We provide empirical evidence for the SSH:
    for each expert, the SAE latents that best predict its activation
    are not semantically related to a single coherent domain (\Cref{sec:spec-hyp:evidence}).
  \item
    We present \emph{RouterInterp}, our SAE-based method, which
    produces accurate explanations of routing decisions,
    achieving a mean explanation score of $0.49$ on gpt-oss-20b and $0.60$ on OLMoE-1B-7B
    (\Cref{sec:routerinterp}).
\end{itemize}

\section{Background}
\label{sec:background}
\subsection{Sparse Autoencoders (SAE)}
\label{sec:background:sae}

A fundamental challenge in interpreting LLM activations is \emph{superposition}:
models encode more features than available dimensions,
resulting in \emph{polysemantic neurons} that activate for multiple unrelated concepts\nopagebreak[4]~\citep{elhage2022superposition}.
Sparse Autoencoders (SAEs) address this by mapping an activation vector
$\vx \in \mathbb{R}^N$ to a sparse latent representation
$\vz \in \mathbb{R}^F$ (where $F > N$).
Ideally, $\vz$ represents a ``disentangling'' of compressed representations into
monosemantic, single-concept latents referred to as \emph{features}
\citep{bricken2023monosemanticity, cunningham2024sparse}.

The SAE architecture consists of an encoder that projects activations into a sparse latent space,
and a decoder that reconstructs the original activation from the sparse latents:
\begin{align*}
  \vz   & = \sigma(\Wenc(\vx-\bpre) + \benc) \\
  \xhat & = \Wdec\vz + \bpre
\end{align*}
where $\Wenc \in \mathbb{R}^{F \times N}$, $\Wdec \in \mathbb{R}^{N \times F}$, and $\sigma$ is a sparsifying activation function.
We follow \citet{gao2025scaling}'s \textbf{Top-K SAE} approach,
where $\sigma = \text{TopK}(\cdot, s)$ retains only the $s$ largest activations
per input and zeros out the rest, ensuring that the latent is sparse\footnote{
  To keep SAE sparsity distinct from MoE routing sparsity, we use $s$ for the former and $k$ for the latter.
}.
The SAE is trained to optimise reconstruction fidelity $\mathcal{L} = \|\vx - \xhat\|_2^2$,
ensuring that the original activation can be accurately
reconstructed from the sparse latent representation $\vz$.

\subsection{Mixture of Experts (MoE)}
\label{sec:background:moe}

In a standard MoE layer, the dense feed-forward network (FFN) is replaced by $E$ parallel expert networks $\{E_i\}_{i=1}^{E}$, each typically an FFN itself.
A learned router network ($\mW_r$) determines which experts process each token.
Given an input token representation $\vx$, the router computes routing logits $h(\vx) = \mW_r \cdot \vx$
and converts them to a probability distribution over experts via softmax:
\begin{align*}
  g_i(\vx) = \frac{\exp(h(\vx)_i)}{\sum_{j=1}^{E} \exp(h(\vx)_j)}
\end{align*}
To enforce sparsity, only the top-$k$ experts with the highest probabilities are selected.
Letting $\mathcal{T}$ denote the set of selected expert indices,
the layer output is computed as the weighted sum of expert outputs:
\begin{align*}
  \vy = \sum_{i \in \mathcal{T}} g_i(\vx) \cdot E_i(\vx)
\end{align*}

A key challenge in MoE training is \emph{load balancing}:
without intervention, models tend to collapse to using only a few experts, leaving others undertrained~\citep{shazeer2017sparse_moe}.
Load balancing is typically addressed through auxiliary losses that encourage
uniform expert utilisation \citep{fedus2022moe_review}.

\section{Specialisation Hypotheses}
\label{sec:specialisation_hypotheses}
\begin{figure}[t!]
  \centering
  \includegraphics[width=0.9\linewidth]{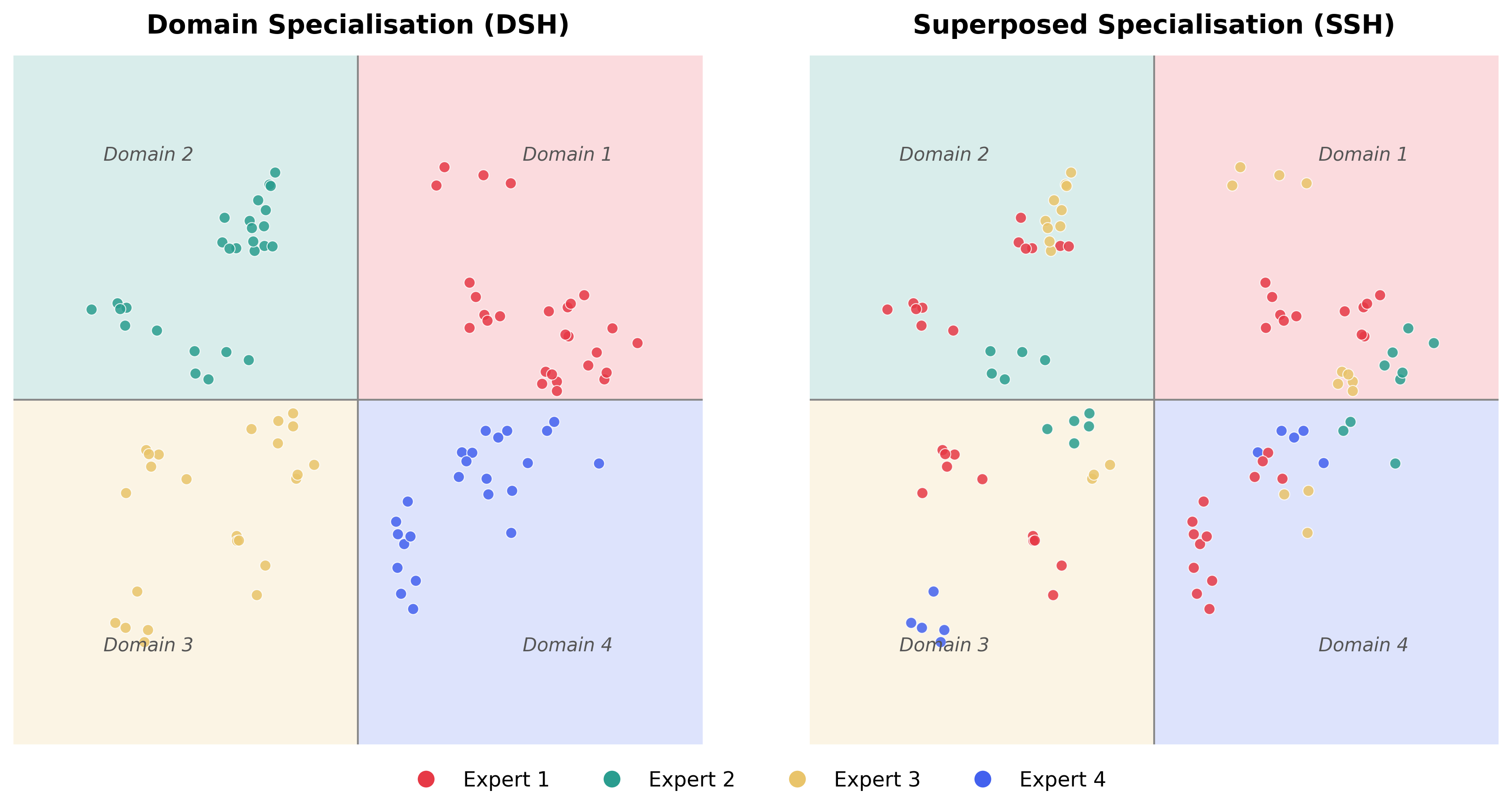}
  \caption{
    Illustration of two alternative hypotheses for the expert specialisation in MoE models.
    \textbf{Left:} Under the Domain Specialisation Hypothesis, examples (points) from the same semantic domain (coloured region)
    are routed to the same expert, so each expert specialises in a single coherent domain.
    \textbf{Right:} Under the Superposed Specialisation Hypothesis, examples routed to the same expert
    are scattered across different domains.
    Here the clusters of adjacent points routed to the same expert
    represent \emph{micro-domains} (fine-grained features of a particular domain)
    and each expert specialises in a disjoint union of them.
    We suggest that the SSH is a more accurate description
    of expert specialisation in practice.
  }
  \label{fig:specialisation_hypotheses}
\end{figure}

In \Cref{sec:intro}, we introduced hypotheses to explain
why MoE models are so effective (\Cref{fig:specialisation_hypotheses}):
the \emph{Domain Specialisation Hypothesis (DSH)} and the
\emph{Superposed Specialisation Hypothesis (SSH)}.
These hypotheses can be understood as claims about whether
expert routing is \emph{monosemantic} (can be explained by
a single concept or domain) or \emph{polysemantic}
(can only be explained by a collection of multiple concepts
or domains which are relevant in different contexts).
The idea of superposition is inspired by \citet{elhage2022superposition},
who describe a theory of the superposition of neurons.\footnote{
  \citet{arora_2018_polysemy} and \citet{goh2016decoding_thought_vector}
  both find evidence for superposition-like phenomena and
  utilise this structure for interpretability similarly to our work.
}
We extend this idea to the superposition of \emph{experts},
arguing that each expert may be specialised in a disjoint collection
of domains.

In this section, we formalise the two hypotheses (DSH and SSH),
state their predictions,
give theoretical motivations for why SSH might hold in practice,
and present experiments that distinguish the two hypotheses.

\subsection{Alternative Specialisation Hypotheses}
\label{sec:spec-hyp:alternative}

Suppose that we have an MoE layer with $E$ experts.
Suppose also that we have a set of $D$ micro-domains occurring
in a corpus $\mathcal{C}$, denoted $\mathcal{D} = \{d_1, \ldots, d_D\}$.
Here, the domains could be topics, genres, or other semantic or syntactic categories
in the data or could alternatively represent sections of the input space that
require the same processing for the next layer of the model.

Firstly, note that when $D = E$ (i.e., when there are the same number of micro-domains as experts),
then we should expect each expert to specialise in a single micro-domain. This is consistent
with both the DSH and SSH.
Secondly, when $D < E$ (i.e., when there are fewer micro-domains than experts),
then we should expect multiple experts to specialise in the same micro-domain or for there
to be redundant experts which are never routed to.
This is also consistent with both the DSH and SSH.

However, in the more interesting and realistic case of $D > E$,
the two hypotheses make different predictions.
Multiple micro-domains must be routed to the same expert and
so the question is whether each expert specialises in multiple
similar micro-domains (as the DSH would imply) or in multiple dissimilar
micro-domains (in accordance with the SSH).

\begin{itemize}
  \item \textbf{Domain Specialisation Hypothesis (DSH):}
    Under the DSH, similar micro-domains are routed to the same expert
    and so an expert specialises in a semantically coherent domain
    consisting of multiple adjacent micro-domains.
    \emph{Prediction: Expert specialisation can be readily explained
    in terms of a single (monosemantic) concept or domain.}
  \item \textbf{Superposed Specialisation Hypothesis (SSH):}
    Under the SSH, dissimilar micro-domains are routed to the same expert
    and so an expert specialises in a collection of semantically
    disjoint domains.
    \emph{Prediction: Expert specialisation can only be explained
    in terms of multiple (polysemantic) concepts or domains.}
\end{itemize}

\subsection{Motivation for Superposed Specialisation}
\label{sec:spec-hyp:motivation}

There are two core arguments for why we might expect the Superposed Specialisation
Hypothesis to be a more accurate description of expert specialisation.

\paragraph{Argument 1 --- Interference Minimisation.}
\citet{elhage2022superposition} argue that superposition is an effective
strategy for neural networks because two sparse features that are not frequently
co-activated can relatively unambiguously share the same set of neurons.
Sparsity, and in particular, low co-activation, reduces the interference cost
of superposition and so allows the model to represent more features
than the number of available neurons.
They find that correlated features tend to be represented orthogonally
because otherwise this would lead to high levels of interference noise
that would make it difficult for the model to disentangle and use these features.

Analogously, we argue that the router is incentivised to assign
dissimilar domains to the same expert so that the
expert can perform Computation in Superposition
\citep{hanni2024comp_in_sup, linsefors2025circuits_in_superposition2,
linsefors2025rotations_in_superposition, newgas2025compressed_computation}.
In other words, a single expert can have multiple disjoint transforms that it
applies depending on the input domain.

\paragraph{Argument 2 --- Load Balancing.}
In MoE training, it is common to include a load-balancing loss
that encourages the expert router to distribute the incoming tokens
evenly across all experts to avoid the under-utilisation of some experts
\citep{shazeer2017sparse_moe,fedus2022moe_review}.
If models are additionally trained with relatively small batch sizes,\footnote{
  OLMoE uses 4M-token batches, about $8\times10^{-7}$ of its training data \citep{muennighoff2025olmoe}; the gpt-oss batch size is unreported.
}
then this can have the unintended effect of encouraging the expert router to
route dissimilar inputs to the same expert.
We can see this because a single sequence is likely to contain mainly
tokens that are from the same macro-domain.
Hence, if the batch size is small, then the batch will not contain very
many different macro-domains; however, the load-balancing loss will
encourage the router to distribute tokens from these few macro-domains
across all of the experts.
This necessarily requires that some tokens from the same macro-domain
will be routed to different experts---in opposition to the DSH,
where we would expect all tokens from the same macro-domain
to be routed to the same expert.

\subsection{Empirical Evidence for Superposed Specialisation}
\label{sec:spec-hyp:evidence}

If the SSH is true, then we should expect that each expert
can be activated by a range of disparate features and hence that
it is generally not possible to explain an expert's routing
behaviour in terms of a single coherent domain.
We test this by treating each expert's top-$n$ most predictive SAE latents
$\sF_i$ (defined in \Cref{sec:methods:feature-selection}) as proxies for the micro-domains
routed to $E_i$ and asking whether
those micro-domains are mutually similar (as predicted by the DSH) or
mutually dissimilar (as predicted by the SSH).

\paragraph{Per-expert micro-domain similarity.}
Let $\text{sim}(f_j, f_k)$ be a similarity measure between
two latents, which represent two micro-domains.
The hypotheses translate directly into predictions about how the latents
in $\sF_i$ behave under $\text{sim}$:
under the DSH the $n$ latents in $\sF_i$ are mutually similar,
and under the SSH they are mutually dissimilar.

To quantify the semantic diversity within $E_i$,
we define $G(E_i)$ as
the number of semantically distinct clusters that $E_i$'s top-$n$ latents fall into.
We then generate a natural-language explanation of each latent from its most activating examples
and compute a clustering of these explanations.\footnote{
  Explanations are embedded with \texttt{all-MiniLM-L6-v2}~\citep{reimers-2019-sentence-bert} and clustered by $k$-means. The silhouette coefficient selects $K{\approx}50$~\citep{rousseeuw1987silhouettes,pedregosa2011scikit}.
}
$G(E_i)$ is the number of distinct clusters occupied by expert $E_i$'s $n$ latents.
The DSH and SSH make opposite predictions for $G(E_i)$:
the DSH predicts $G(E_i){=}1$ (all of $E_i$'s latents fall in a single
cluster, representing one coherent micro-domain), whereas the SSH
predicts $G(E_i){\to}n$\footnote{
  High $G$ could also mean \emph{inefficient} routing (fails to exploit structure)
  or \emph{inexplicable} routing (not semantically meaningful).
  MoE performance provides evidence against the former;
  human-interpretable RouterInterp structure provides evidence against the latter
  (\Cref{sec:routerinterp} and \Cref{sec:appendix:human_interpretability}).
}
(latents tend to occupy distinct clusters, each representing
a semantically distinct micro-domain).

\begin{figure}[t!]
  \centering
  \includegraphics[width=0.9\linewidth]{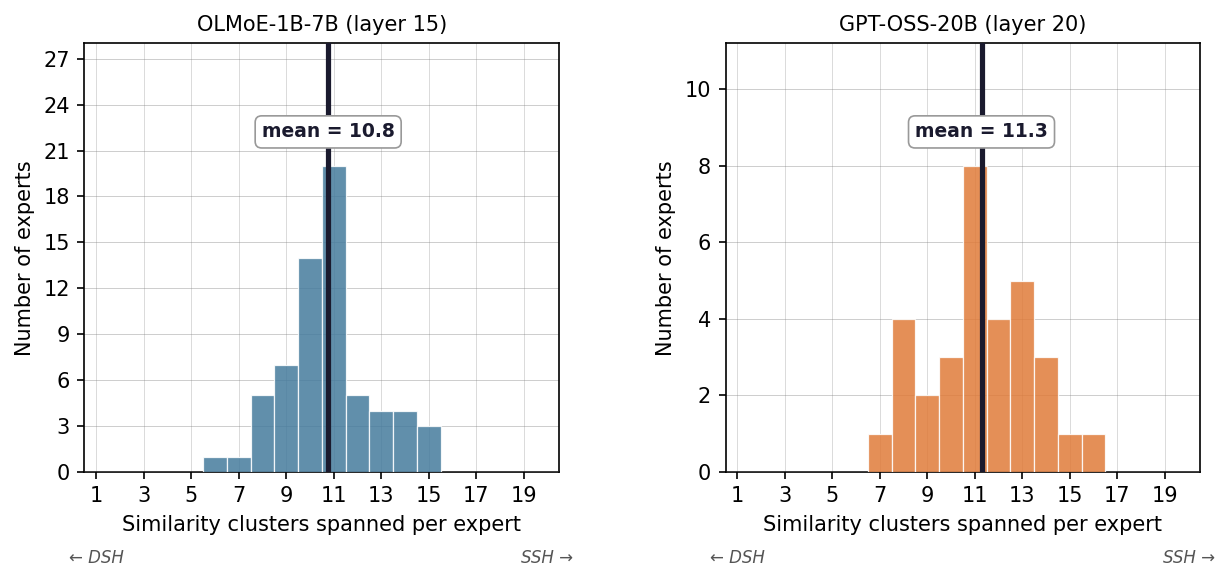}
  \caption{
    \textbf{Each expert covers a semantically diverse range of micro-domains:
    the latents that best predict its activation are spread across many distinct semantic clusters.}
    For each expert we take its top $n{=}20$ predictive SAE latents $\sF_i$
    and cluster them semantically by their explanations.
    The DSH predicts one shared cluster; the SSH predicts a separate cluster per latent.
    We observe that experts' latents are grouped into far more than one cluster:
    10.8 clusters on average for OLMoE-1B-7B (layer 15) and 11.3 for gpt-oss-20b (layer 20).
  }
  \label{fig:ssh_dsh_summary}
\end{figure}

\Cref{fig:ssh_dsh_summary} shows the distribution of $G(E_i)$ across experts
at the deepest tested layer of each model.
The empirical $G$ values are far from the value of $1$ predicted by the DSH
in both models:
mean $G$ is $11.3$ on gpt-oss-20b (layer~20) and $10.8$ on OLMoE-1B-7B (layer~15),
out of an upper bound of $n{=}20$.
The observed values lie far above the DSH prediction of $G{=}1$, 
yet slightly below the random\footnote{
  Random cluster assignment would be about $16.6$ clusters for $20$ latents among $50$ equal-sized clusters.
} cluster assignment.
This suggests that experts retain some thematic coherence
(their predictive latents are more related than arbitrary collections)
while still spanning many disjoint micro-domains.
We take this as evidence for the SSH over the DSH.

We also find that routing-prediction F1 generally degrades as fewer SAE features are
kept active per token (\Cref{sec:appendix:num_features_ablation}),
and that routing across Pile subsets stays near corpus-proportional
(\Cref{sec:appendix:pile_domains}),
suggesting that routing depends on many features jointly
rather than on any single domain-level signal.

\section{RouterInterp}
\label{sec:routerinterp}
In \Cref{sec:specialisation_hypotheses},
we provided evidence that expert routing is well described by the SSH.
Based on this insight, we propose \textbf{RouterInterp},
our method for generating natural language explanations
of expert routing using SAE features associated with each expert.

\begin{figure}[t]
  \centering
  \includegraphics[width=0.8\linewidth]{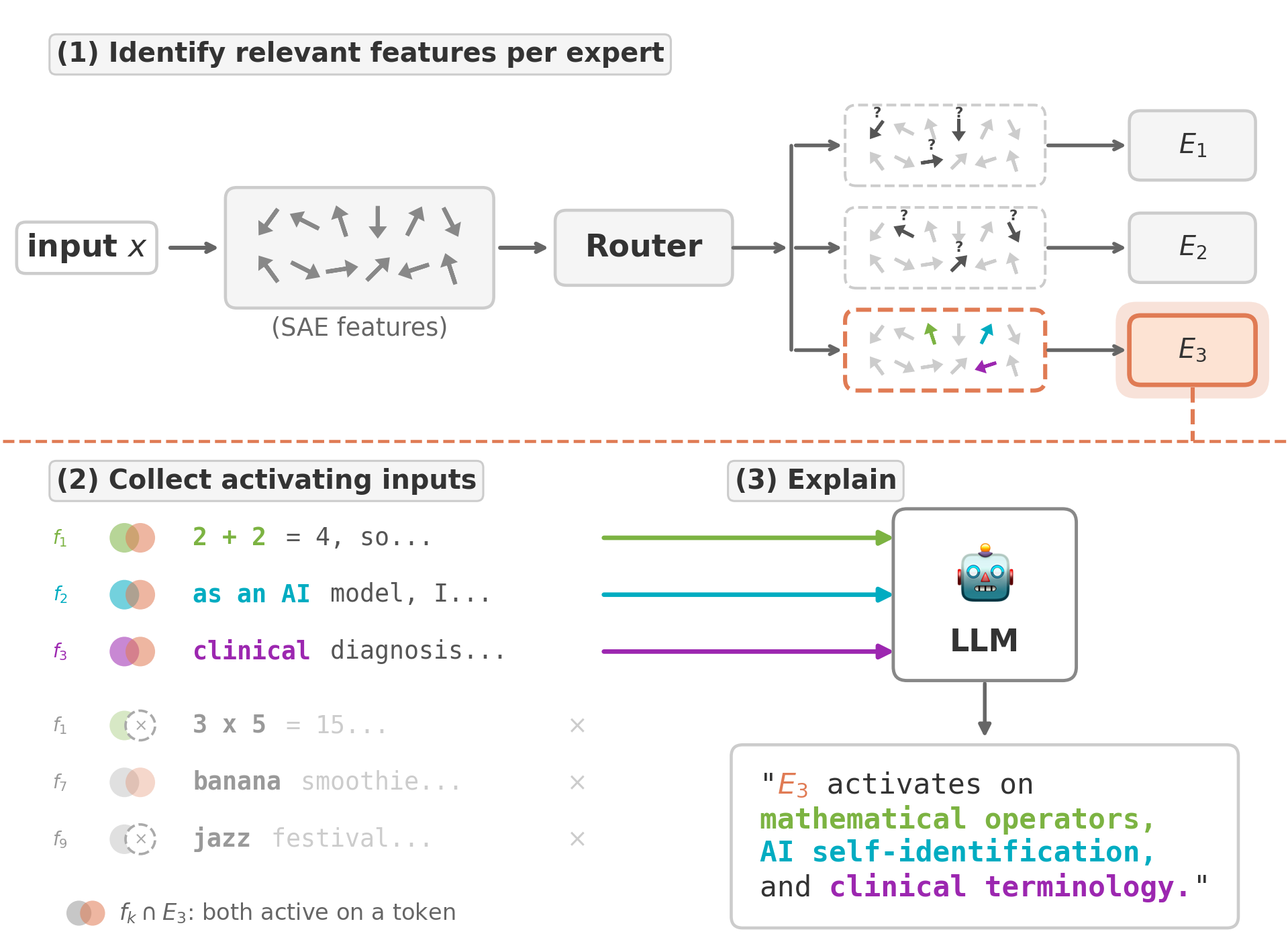}
  \caption{
    RouterInterp explains routing decisions as a combination of interpretable features.
    Sparse autoencoder (SAE) latents, sparse feature representations extracted from model activations, pick out directions in activation space that correspond to interpretable features.
    RouterInterp \emph{identifies} SAE latents that are most predictive of routing,
    \emph{collects} inputs where these features and the expert are jointly active,
    and \emph{explains} the expert in natural language.
    While no single monosemantic concept captures an expert's behaviour,
    aggregating multiple features makes routing understandable.
  }
  \label{fig:hook_figure}
\end{figure}

\subsection{Method}
\label{sec:methods:routerinterp}

RouterInterp operates in three stages (\Cref{fig:hook_figure}):
(1) \emph{identifying} the SAE features most relevant to each expert's routing behaviour,
(2) \emph{collecting examples} where each selected feature and the expert are active together,
and (3) \emph{explaining} the expert with a language model prompted on these activations.

\paragraph{Identifying Features.}
\label{sec:methods:feature-selection}
We select the SAE features most useful for each expert
by approximating the effect of ablating each feature on
whether the expert is selected (\Cref{sec:appendix:feature_selection:method}).

For each expert $E_i$, we take the top-$n$ features by this approximated ablation effect,
denoting this set $\sF_i$;
each feature is assigned to at most one expert
so that the selected sets isolate the routing signals that distinguish experts from one another.
\footnote{
  We evaluate the effect of feature set size $|\sF_i|$ on routing prediction quality
  and compare against alternative selection methods in
  \Cref{sec:appendix:feature_selection:criterion}.
}.

\paragraph{Collecting Activations.}
\label{sec:methods:explanation-generation}
For each selected feature $f \in \sF_i$ of expert $E_i$,
we collect context windows centred on tokens where $f$ is active
and form two sets:
the positive set $\mathcal{P}_{i,f}$, where $f$ is active and the token is routed to $E_i$,
and the negative set $\mathcal{N}_{i,f}$, where $f$ is active but the token is \emph{not} routed to $E_i$.

\paragraph{Explaining.}
Given the feature set $\sF_i$ and the associated example sets
$\{(\mathcal{P}_{i,f}, \mathcal{N}_{i,f})\}_{f \in \sF_i}$,
we prompt a language model (the \emph{explainer}) with these examples
grouped by feature:
for each $f \in \sF_i$, the explainer sees both $\mathcal{P}_{i,f}$ and $\mathcal{N}_{i,f}$,
ranked by feature activation strength, with activating tokens highlighted.
This positive--negative contrast within each group isolates the routing-relevant context.
It helps the explainer identify what makes the expert active \emph{given} the feature.
We ask the explainer to synthesise these groups into a single prose description of when expert $E_i$ is selected,
merging overlapping patterns across features while keeping genuinely distinct activation cases separate.
Prompts are provided in \Cref{sec:appendix:prompts}.

\subsection{Evaluation}
\label{sec:methods:evaluation}

\paragraph{Scoring.}
\label{sec:methods:scoring}

To evaluate whether our expert explanations accurately capture routing behaviour,
we adapt the AutoInterp \emph{Detection} scoring setup from \citet{paulo2025automatically}.
In AutoInterp, another language model (the \emph{scorer}) uses the explanation to predict
expert activation on held-out examples.
We design an analogous scoring approach for expert routing to
show that not only can we mathematically describe routing behaviour but also that
we can do so in natural language explanations that are human-interpretable.

For each expert $E_i$, we build a held-out evaluation set of context windows.
\emph{Positive} windows contain at least one token routed to $E_i$;
\emph{negative} windows are sampled uniformly from windows in which $E_i$ is never selected.
We present the scorer with the expert explanation and each window,
highlighting the tokens routed to $E_i$ on positives
and randomly chosen tokens on negatives.
The number of highlights for each negative window is sampled from the empirical distribution of highlight counts on positives.
The scorer judges whether at least one highlighted token, in its surrounding context, matches the explanation.
We threshold these judgments to a binary prediction and compare against the ground-truth routing label.
We report F1 score of this binary classification task,
measuring how well the explanation captures the expert's routing behaviour.
Scorer details are given in \Cref{sec:appendix:prompts}.
We also run a small-scale human evaluation to validate whether the LLM scorer is reliable:
\Cref{sec:appendix:human_interpretability} shows that human--LLM agreement is comparable to human--human agreement,
supporting its use as a proxy for human interpretability judgments.

\paragraph{Baselines.}
We compare RouterInterp against prior work on interpreting expert routing.
\citet{jiang2024mixtral,zoph2022st_moe} analyse routing via token co-occurrence statistics.
We formalise their approach as \textbf{Unigram Lookup}:
for each expert $E_i$, we record the top-$100$ tokens that most frequently co-occur with its activation
and predict positive if any highlighted token belongs to this set.
For lookup methods, we do not use the scorer LLM
and instead produce predictions directly from their frequency tables.
We also compare against the expert-level automatic interpretability pipeline of~\citet{herbst2026expert},
which we refer to as \textbf{Expert Impact AutoInterp}.
Their approach prompts a language model to generate an expert explanation based on:
\emph{(a)} passages that contain tokens
where the target expert $E_i$ writes most strongly to the residual stream,
sorted by $g_i(\vx)\,\lVert E_i(\vx)\rVert_2$, and
\emph{(b)} top vocabulary tokens promoted by the expert at peak activation,
recovered via Logit Lens~\citep{nostalgebraist2020interpreting}.

\paragraph{Ablations.}
To understand which components contribute to RouterInterp's improvement over baselines,
we evaluate a series of ablations that progressively enrich the explainer's context.
\textbf{Bigram Lookup} extends Unigram Lookup with local token context (current and previous token).
\textbf{Unigram AutoInterp} and \textbf{Bigram AutoInterp} use the same token statistics
but prompt a language model to summarise the frequent (bi)grams into a natural language description.
\textbf{Expert Activations AutoInterp} replaces n-gram summaries with full activating windows:
for each expert $E_i$, we collect passages that contain tokens routed to $E_i$
(ranked by router scores $g_i(\vx)$)
and prompt a language model to generate an expert-level explanation from those contexts.
This isolates the gain from providing the LLM with whole-passage context rather than local (bi)grams,
still without SAE decomposition (RouterInterp) or impact-based ranking (Expert Impact AutoInterp).
We report all prompts used for explanation generation in \Cref{sec:appendix:prompts}.

\subsection{Experimental Setup}
\label{sec:setup}

We evaluate on two MoE architectures: OLMoE-1B-7B~\citep{muennighoff2025olmoe} and gpt-oss-20b~\citep{agarwal2025gpt_oss}.
For training OLMoE SAEs, we use OLMoE-mix-0924~\citep{muennighoff2025olmoe}.
For routing analysis and explanation scoring, we collect activations using the Pile
dataset~\citep{gao2020pile}.
For OLMoE, we train Top-K SAEs on 100M tokens of activations sampled from
layers 3, 7, 11, and 15 (out of OLMoE's 16 layers),
with per-token latent sparsity $s=32$ and 32,768 features.
For gpt-oss (24 layers), we use trained BatchTopK SAEs%
\footnote{\url{https://huggingface.co/andyrdt/saes-gpt-oss-20b}}~\citep{lin2025neuronpedia_fall}
on layers 4, 8, 12, 16, and 20, with 131,072 features and sparsity $s \in \{64, 128\}$.
We report SAE training metrics in \Cref{sec:appendix:sae_metrics}.

We generate RouterInterp explanations across layers using the top-45 features per expert
selected by gradient-based attribution.
We use the Delphi library \citep{paulo2025automatically}
for explanation generation and scoring,
with Claude Sonnet~5~\citep{anthropic2026claude_sonnet_5} as the explainer
and GPT-5.6 Luna~\citep{openai2026gpt56} as the scorer.
The explanation score is computed over all experts per layer,
with 15 positive and 85 negative examples per expert,
so that the positive rate ($15\%$) approximates the fraction of tokens
any single expert receives in the model
($k/E \approx 12.5\%$ for gpt-oss with $k{=}4$, $E{=}32$
and for OLMoE with $k{=}8$, $E{=}64$).

\subsection{Results}
\label{sec:results:interpretability}

\begin{figure}[H]
  \centering
  \includegraphics[width=0.9\linewidth]{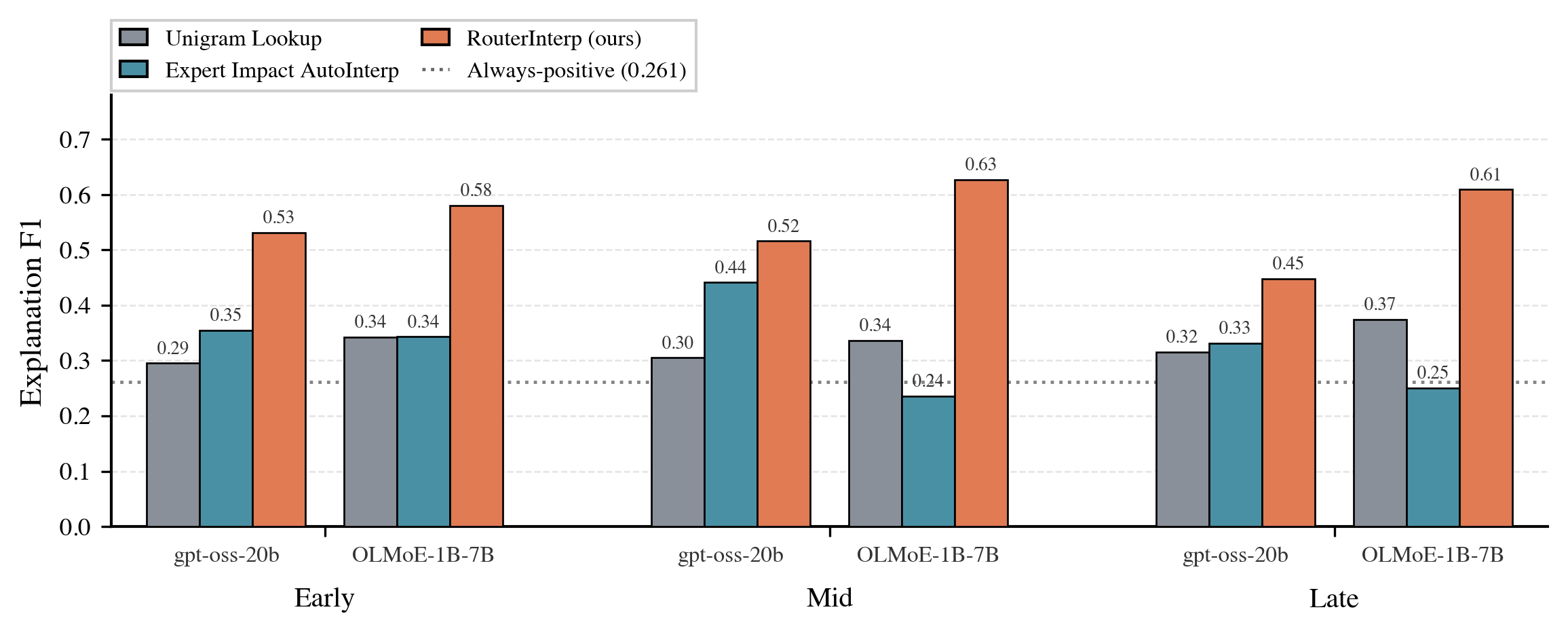}
  \caption{
    Leveraging the SSH,
    RouterInterp generates expert explanations from examples where SAE features and experts are jointly active.
    It outperforms both a method based on token-statistics~\citep{jiang2024mixtral,zoph2022st_moe}
    and Expert Impact AutoInterp~\citep{herbst2026expert},
    which generates explanations from activating windows ranked by expert impact on the residual stream,
    at every model depth
    (${\sim}65\%$ / ${\sim}76\%$ mean F1 over Unigram Lookup on gpt-oss-20b / OLMoE-1B-7B;
    ${\sim}28\%$ over Expert Impact AutoInterp on gpt-oss-20b).
    Shown for layers 4, 16, 20 of gpt-oss-20b
    and layers 3, 11, 15 of OLMoE-1B-7B.
  }
  \label{fig:main_results}
\end{figure}

\Cref{fig:main_results} reports explanation scores for each method across layers of gpt-oss-20b and OLMoE-1B-7B.
RouterInterp attains mean F1 of $0.492$ ($s{=}128$) on gpt-oss-20b and $0.602$ on OLMoE-1B-7B,
versus $0.383$ and $0.305$ for Expert Impact AutoInterp
and $0.299$ and $0.342$ for Unigram Lookup respectively,
winning at every layer on both models.

We attribute RouterInterp's advantage to the fact that it generates explanations from activations of SAE features.
Under the SSH,
examples selected along a single dimension (e.g. token frequency, activation strength, or residual-stream impact)
cannot cover an expert's disjoint micro-domains.
The resulting explanation either collapses onto the expert's dominant activation pattern
or stays too generic to characterise its routing.
RouterInterp instead enumerates the expert's micro-domains via its top-$45$ attribution-selected features\footnote{
  45 features and attribution ranking outperform smaller feature sets and other feature selection methods,
  see \Cref{sec:appendix:feature_selection} for ablations.
}
and shows routed versus non-routed windows per feature,
so its explanations cover most of the expert's micro-domains at once.

\begin{table}[H]
  \centering
  \begin{tabular}{lccccc}
    \toprule
    & \textbf{Layer 4} & \textbf{Layer 8} & \textbf{Layer 12} & \textbf{Layer 16} & \textbf{Layer 20} \\
    \midrule
    Unigram Lookup            & 0.295 & 0.300 & 0.278 & 0.305 & 0.315 \\
    Bigram Lookup             & 0.309 & 0.320 & 0.283 & 0.320 & 0.331 \\
    Unigram AutoInterp        & 0.332 & 0.406 & 0.320 & 0.356 & 0.348 \\
    Bigram AutoInterp         & 0.429 & 0.453 & 0.385 & 0.407 & 0.411 \\
    Expert Activations AutoInterp & 0.415 & 0.475 & 0.345 & 0.425 & 0.425 \\
    \midrule
    RouterInterp ($s{=}64$)
                              & 0.490 & \textbf{0.542} & \textbf{0.465} & 0.509 & \textbf{0.470} \\
    RouterInterp ($s{=}128$)
                              & \textbf{0.531} & 0.509 & 0.456 & \textbf{0.516} & 0.448 \\
    \bottomrule
  \end{tabular}
  \caption{
    Expert explanations improve when an LLM summarises evidence rather than using raw lookup tables,
    but n-grams and expert activations still cannot cover an expert's disjoint micro-domains
    (both $0.417$ mean F1).
    RouterInterp uses SAE features to enumerate those micro-domains and
    gains $+18\%$ over the best of these LLM-based explainers.
    RouterInterp's success is also robust to SAE sparsity ($0.495$ for $s{=}64$ vs.\ $0.492$ for $s{=}128$).
    Results are reported on gpt-oss-20b.
  }
  \label{tab:expert_explanations_ablations}
\end{table}

The ablations in \Cref{tab:expert_explanations_ablations} confirm that this gain
comes from the SAE features rather than from the LLM explainer itself.
Verbalising token statistics with an LLM improves explanations only modestly
(from 0.30 / 0.31 to 0.35 / 0.42 mean F1),
and replacing token lists with full activating passages does not help either:
Expert Activations AutoInterp performs on par with Bigram AutoInterp ($0.417$ mean F1 on gpt-oss-20b).
The two ablations fail in the two ways the SSH predicts:
bigram descriptions are narrow
(higher precision, lower recall: P $0.39$ / R $0.59$),
while full-window descriptions are broad
(lower precision, higher recall: P $0.30$ / R $0.81$).
Only when the evidence is grouped by SAE feature do precision and recall improve together
(P $0.47$ / R $0.64$; $0.495$ mean F1, $s{=}64$),
confirming SAE feature decomposition as the key ingredient of RouterInterp.

Expert Impact AutoInterp~\citep{herbst2026expert} also uses full activating examples,
but with a different example ranking ($g_i(\vx)\,\lVert E_i(\vx)\rVert_2$) plus Logit Lens tokens.
This approach does not beat simply showing activating windows ($0.383$ vs.\ $0.417$ mean F1 on gpt-oss-20b)
and is the most volatile method across layers and models.
On OLMoE-1B-7B it even falls below Unigram Lookup ($0.305$ vs.\ $0.342$):
we found that for $19\%$ of experts the descriptions are overly narrow and score exactly zero.
\citet{herbst2026expert} report F1 ${>}0.8$ for most experts, but in a different evaluation setting:
their evaluation set is drawn from maximum-activation examples with $10$/$10$ positive--negative balance
and their scorer prompt accepts rough hypothesis fit on target tokens,
whereas ours rejects mere token overlap
and requires a contextual match to a concrete activation pattern.
The absolute scores are therefore not directly comparable
between our evaluation and that of \citet{herbst2026expert}.\footnote{
  A trivial classifier that always predicts \emph{active}
  already achieves an F1 of $0.67$ at $10$/$10$ balance,
  versus $0.26$ at our $15\%$ positive rate.
}

RouterInterp's advantage over baselines is much larger on OLMoE-1B-7B,
where it scores $0.58$--$0.63$ across layers, than on gpt-oss-20b.
We believe this is because the smaller model's routing is easier to characterise,
and because its SAEs are smaller and reconstruct better
($32$k latents, FVU $0.007$--$0.218$, versus $131$k latents for gpt-oss).
The same dependence on SAE quality is visible within gpt-oss-20b itself:
layer~16 has the worst SAE reconstruction (FVU $0.19$--$0.23$)
and the weakest routing prediction,
and is the only layer where a matched-sparsity neuron probe outperforms
the SAE basis (\Cref{sec:appendix:routing_prediction_layers}).
We discuss the effect of the underlying SAE further in \Cref{sec:appendix:sae_metrics}
and show example explanations in \Cref{sec:appendix:expert_explanations}.

\section{Discussion}
\label{sec:discussion}
Our results validate the Superposed Specialisation Hypothesis (SSH)
and confirm that expert routing is highly polysemantic.
The set of features most aligned with any single expert often spans disjoint concepts,
meaning experts do not specialise in a single monosemantic domain.
This finding is supported by our automated expert interpretations,
which frequently describe experts as activating on logical disjunctions of unrelated features\footnote{
  Examples include an expert activating on both ``C/C++ preprocessor directives'' and
  ``spices and herbs'' or ``medical conditions'' and ``electronic components''.
}.
Superposed Specialisation suggests why previous work
(that implicitly assumed monosemantic experts as in the DSH)
was unable to successfully explain expert routing behaviour.

In recent work, there have been examples of apparently
interpretable specialisation in expert routing:
\citet{lasby2026reap} find that for specific tasks like code generation
they can prune half of the experts and retain most of the model quality
for that task;
\citet{bandarkar2025multilingual_routing} show that some multilingual models
contain experts that fire preferentially on particular languages;
and \citet{fayyaz2026steermoe} demonstrate that
ablating a small set of experts can reliably toggle
safety refusals.
As it relates to the SSH,
in each of these cases of apparent domain specialisation the result shows that an expert
\emph{participates} in a language, task, or behaviour;
none of the results show that this is the \emph{only} thing the expert does.
Our experiments in \Cref{sec:spec-hyp:evidence} show that each
expert covers many disjoint semantic roles, far from a single coherent domain.
An expert that fires often on Spanish text, for example, will still have many other roles,
and Spanish will be only a small fraction of what it does over a broader distribution.

One remaining question about how experts specialise is whether
experts aggregate tokens based on
\emph{input similarity} (grouping tokens that contain similar semantic features)
or \emph{functional similarity} (grouping tokens that require similar downstream transformations).
We suggest that expert specialisation based on functional similarity may be
a productive way to understand routing,
with each expert performing Computation in Superposition
\citep{hanni2024comp_in_sup, newgas2025compressed_computation}.
In this way, experts may group seemingly unrelated inputs that
require a set of disjoint computational operations to be applied in the current layer.
We would be excited about future work testing the functional routing hypothesis,
as understanding the principles guiding expert specialisation could inform
the design of more efficient and interpretable MoE architectures.

Our analysis of specialisation started with the observation that there
are more micro-domains than experts and
specialisation can then be understood as how the micro-domains are clustered together.
Other approaches to building or interpreting MoE models may also get
around this problem.
For building more interpretable MoE models,
\citet{park2025monet} show that with a much larger number of experts
expert routing becomes increasingly monosemantic and interpretable.
For interpreting existing MoE models,
despite the fact that there are few experts,
there are combinatorially many routing collections
(i.e.\ the top-k experts for a given token)
or routing paths (i.e.\ the sequence of experts a token passes through across layers).
It may be that interpreting routing collections or paths provides a more
tractable way to understand routing decisions.

We also note that our results provide an example of a success case for
SAE-based methods.
There has been some debate in the interpretability literature about the effectiveness of
SAEs and dictionary learning recently.
\citet{movva2025saes_for_hypothesis_generation}, \citet{jiang2025saes_for_data_analysis}
and \citet{lindsey2025biology_of_llms} find SAEs useful for
generating hypotheses and understanding model behaviour;
whereas \citet{kantamneni2025saes_for_probing} and \citet{wu2025axbench} find that
SAEs are not optimal in their probing use cases.
We see our success using SAEs in the context of \citet{peng2025saes_for_discovery_not_action}
who suggest that SAEs are useful for discovery of unknowns rather than acting on known concepts.
We also note that in other cases where linearity is not guaranteed,
other unsupervised methods for extracting feature manifolds may be effective
such as in \citet{bhalla2026concept_manifolds}.

\subsection{Limitations}
Our method's effectiveness depends on the quality of both SAEs and the automated explanation pipeline.
Concretely, RouterInterp performance correlates with SAE reconstruction quality across layers:
higher FVU corresponds to weaker routing prediction
(see \Cref{sec:appendix:sae_metrics}).

When aggregating the contributions of multiple features, RouterInterp produces
natural language explanations that increase in size with more features.
Given the length considerations for an increasing number of features,
methods which further compress the explanations to be more concise might
provide useful optimisations to our method.
Similarly, interactive presentation formats,
such as feature dashboards on Neuronpedia \citep{neuronpedia},
may better enable exploration of the feature-expert relationships that drive routing decisions.

Finally, our evaluation was performed on two models (OLMoE-1B-7B and gpt-oss-20b) in the 1B--20B parameter range,
and we would be excited about future work scaling RouterInterp
to frontier architectures with more parameters,
shared experts and different routing methods
like Expert Choice routing.

\section{Related Work}
\label{sec:related_work}
\citet{shazeer2017sparse_moe} designed the modern Sparse MoE layer to improve
the efficiency and scalability of very large neural networks.
Other researchers, however, hoped that the specialisation routing provided
might also extend some interpretability benefits \citep{jacobs1991original_moe, fedus2022moe_review}.
However, \citet{jiang2024mixtral} find it difficult to obtain clean interpretability results
with their MoE model Mixtral, remarking ``Surprisingly, we do not observe obvious
patterns in the assignment of experts based on the topic [of the input text].''
\citet{lewis2021base_moe} and \citet{zoph2022st_moe} analyse
the unigram patterns of routing and find that
the observed level of specialisation varies dramatically making interpretability difficult.
\citet{tigges2025goodfire} report some success with unigram analysis of a few experts such as
a `business' expert.
We show, however, that by using SAE features as a basis for explanation
rather than tokens, we are able to robustly achieve more accurate explanations.

\citet{yang2025interpretable_moe} suggest that with very large numbers of experts,
routing can be made somewhat interpretable.
However, their setting requires more experts than is compute optimal given scaling laws for
sparse models \citep{apple2025parameters_vs_flops, krajewski2024scaling_laws_moe}
or more experts than is typical in high performing open source MoE models
\citep{liu2024deepseek_v3, yang2025qwen3technicalreport, agarwal2025gpt_oss, muennighoff2025olmoe}.

\citet{chaudhari2025superposition_in_moes} provide an alternative model for understanding
the phenomena of superposition \citep{elhage2022superposition} in MoE models.
\citet{chaudhari2025superposition_in_moes} find that individual experts exhibit
greater monosemanticity than equivalent dense models (in a toy setting).
Crucially, their analysis measures monosemanticity at the level of
\emph{expert weight matrices}, not at the level of \emph{routing decisions},
which we focus on.
We argue that even if individual experts represent their assigned features
monosemantically, the routing decision itself may be polysemantic.
The router partitions the input space such that features assigned to the same
expert only compete with each other for representational capacity
\citep{chaudhari2025superposition_in_moes}.
This creates an incentive to route \emph{dissimilar} domains to the same expert:
micro-domains that rarely co-occur can share an expert with little interference,
enabling the expert to effectively perform Computation in Superposition
\citep{hanni2024comp_in_sup} as we detail in \Cref{sec:specialisation_hypotheses}\footnote{
  We provide an extended review of additional related
  works in \Cref{sec:extended_related_work}.
}.

\section{Conclusion}
\label{sec:conclusion}
We introduced \emph{RouterInterp}, a method that interprets MoE routing decisions
by identifying sparse autoencoder features most predictive of expert selection
and aggregating their explanations into unified natural language descriptions.
RouterInterp produces descriptions that capture not just \emph{which} tokens route to an
expert, but \emph{why}.
Our results confirm that SAE features provide
both predictive accuracy for routing and semantic interpretability:
on gpt-oss-20b and OLMoE-1B-7B, RouterInterp's natural language explanations
achieve a $0.49$ / $0.60$ explanation score,
outperforming both explanations generated directly
from expert-activating text without SAE decomposition ($0.42$ on gpt-oss-20b)
and unigram-based explanations ($0.30$ / $0.34$).

RouterInterp emerged from formalising two alternative specialisation hypotheses:
the \emph{Domain Specialisation Hypothesis} (DSH), which posits that experts specialise in
semantically coherent domains,
and the \emph{Superposed Specialisation Hypothesis} (SSH),
where experts respond to disjoint collections of unrelated micro-domains.
Our experimental results support the SSH and suggest that
experts are highly polysemantic and cover multiple domains.
This is supported by our automated expert interpretations,
which frequently describe experts as activating on logical disjunctions of unrelated features.

Several directions remain open for future investigation.
First, the \emph{mechanisms driving expert specialisation} deserve deeper study:
why do experts converge to their particular feature combinations rather than others, and
do load-balancing losses or model capacity constraints force redundancy across experts?
Understanding whether experts cluster based on input similarity or functional similarity---that is,
whether co-routed tokens are co-routed because they share semantic content or
because they require similar downstream transformations---
could reveal fundamental principles of how modular architectures self-organise.

Second, tracking how routing evolves across \emph{training checkpoints}
could reveal when and how expert specialisation emerges
and how early training shapes routing.
This problem could be attacked similarly to the Developmental Interpretability literature
\citep{wang2024loss_landscape_geometry,kangaslahti2025llm_training_breakthroughs};
for example, \citet{wang2025differentiation_and_specialisation_of_attention} study the specialisation of attention heads over training.

RouterInterp takes a step toward making expert routing a transparent and
controllable component of foundation models.
As MoE architectures continue to power frontier systems,
tools that reveal the logic behind routing decisions will be essential---not only for scientific understanding,
but for ensuring these systems remain aligned with human intent.

\section*{Acknowledgements}
Thanks to Ivaylo Dimitrov for contributions to an earlier version of this project.
Thanks to Cameron Holmes, Kamal Maher, Simon Schrader, and Aran Arslan for helpful comments on
earlier drafts of this work.
Thanks to David Africa, Arathi Mani, Edmund Lau, Sid Black,
Nora Belrose, Gonçalo Paulo, Lucia Quirke, Herbie Bradley,
Andrew Gritsevskiy, Derik Kaufmann, Lucius Bushnaq, Daniel Filan,
Hans Gundlach, Clément Dumas, and Peter Knees for helpful conversations.
We are grateful to the SPAR Programme for additional facilitation of this project.
Thanks to Cameron Holmes and Benjamin Hilton for additional support.
We appreciate the support of Cavendish Labs where some ideas for
this project were conceived.

The authors acknowledge the use of resources provided by the Isambard-AI National AI Research Resource (AIRR).
Isambard-AI is operated by the University of Bristol and is funded by the UK Government’s
Department for Science, Innovation and Technology (DSIT) via UK Research and Innovation;
and the Science and Technology Facilities Council [ST/AIRR/I-A-I/1023].
IL was funded in part by the Austrian Science Fund (FWF) 10.55776/COE12.

\section*{Impact Statement}
We present a new method for interpreting expert routing in
Sparse Mixture-of-Experts (MoE) models.
When combined with existing techniques for interpreting attention, MLPs, and intermediate representations,
this work can help researchers understand the inner workings of neural networks.

We anticipate that the overall effect of this work will be to accelerate progress in mechanistic
interpretability and consequently improve our ability to explain and steer model behaviour.
Though we acknowledge the potential dual use nature of interpretability research
(as with all Machine Learning research),
we expect that the main applications of this work will be in debugging neural networks,
improving the trustworthiness of AI systems,
enabling the evaluation of fairness and bias in model decision-making,
and understanding and mitigating potential risks from AI systems.

\bibliographystyle{icml2026}
\bibliography{references}

\newpage
\appendix
\crefname{appendix}{Appendix}{Appendices}
\Crefname{appendix}{Appendix}{Appendices}
\crefalias{section}{appendix}

\startcontents[appendices]

\section*{Appendix Contents}
\printcontents[appendices]{}{1}{\setcounter{tocdepth}{2}}

\section{Extended Related Work}
\label{sec:extended_related_work}
\subsection{Sparse Feature Decompositions for Model Interpretability}
\label{sec:related_work:sae_interp}

SAEs rely on the Linear Representation Hypothesis
\citep{mikolov2013word2vec, bricken2023monosemanticity, olah2020zoom_in, nanda2023othello_gpt, park2024lrh}:
the hypothesis that features in LMs are represented linearly \citep{bereska2024mech_interp_survey}.
The LRH is often a controversial assumption within interpretability
\citep{smith2024strong_feature_hypothesis,csordas_2024_rnns_use_non_linear_representations}.
However, in our case of interpreting expert routing,
routing is usually performed by a linear map and so only linearly
accessible information can be important for routing.
This makes SAEs uniquely well suited to our problem\footnote{
  Also note that the Tuned Lens \citep{belrose2023tuned_lens}
  also validates that linear probes on intermediate residual
  stream states capture meaningful structure.
}.

SAEs have drawbacks, however.
SAEs can struggle to capture logical hierarchical
structure \citep{costa2025mp_sae};
they do not efficiently capture structures of multi-dimensional features
\citep{engels2025multi_dimensional_features};
they do not generally capture relations across multiple tokens
\citep{lubana2025priors_in_time};
and the structure of Relational Composition \citep{wattenberg2024relational_composition}
remains difficult for SAEs.

Most work interpreting representations to date has focused on the impact of intermediate representations
on the output logits of a forward pass \citep{bricken2023monosemanticity}
or on representations in a subsequent layer \citep{marks2025sparse_feature_circuits}.
We instead focus on how representations impact the behaviour of a subsequent routing layer.

Inspired by the apparent success of SAE-based explanations,
\citet{sharkey2024sparsify} describes a framework for the continual improvement of MI explanations.
The framework’s three stages are:
(1) \emph{mathematical description}
(breaking down the neural network into functional parts\footnote{
    where here the parts could be in terms of representations \citep{bricken2023monosemanticity}
    or computations \citep{braun2025spd}
}),
(2) \emph{semantic description}
(labelling each functional part in a way that is understandable to humans\footnote{
    Note that implicitly this formulation requires the Independent Additivity Principle
    \citep{ayonrinde2024mdl_saes} - if multiple parts are relevant at one time,
    then it must be the case that we can understand the whole simply from understanding the constituent parts
})
and (3) \emph{validation} (using the semantic description to make predictions about model behaviour and evaluating
these predictions).
Here our combination of the trained SAEs and our Ansatz (conjecture) that router weights are linear combinations
of weights provide our mathematical description of routing. We leverage AutoInterp
\citep{paulo2025automatically} to aid with
semantic description and evaluation of the functional parts and the empirical success of this approach
provides us with evidence for the Ansatz.

\subsection{Adaptive Computation}

Sparse MoE models are a type of Adaptive Computation model
\citep{graves2016act, ayonrinde_2023_awesome_adaptive_computation}\footnote{
  Also known as Conditional or Dynamic Computation models \citep{han2021dynamic}
}.
Adaptive Computation models are typically either more FLOP or parameter efficient than dense models
as they either use a subset of their parameters
(for example sparse MoE models or Early Exit models \citep{banino2021pondernet})
or reuse parameters (for example looping models
\citep{dehghani2018universal_transformers, yang2024looped_transformers}) respectively.
There has been little work on the interpretability of Adaptive Computation models and
while our work focuses on interpreting routing in the sparse MoE layer, we would be
excited about work generalising this approach to routing in early-exit models
like Mixture of Depths \citep{raposo2024mixture_of_depths}.

\subsection{Modularity in Neural Networks}

Many brain-inspired neural network architectures employ modularity
(the organisation of a system into functional, sparsely connected subunits)
as a core component \citep{pfeiffer2023modular_deep_learning},
as modularity is believed to be one of the key
properties that makes brains efficient and effective \citep{clune2013evolution_of_modularity}.

We might hypothesise that because neural networks generalise so well
and have much lower effective dimensionality than the model dimension would suggest \citep{lau2025llc},
we should expect the models to contain intrinsic modularity.
That is to say that even though NNs look fully connected and highly entangled,
perhaps there is some way of viewing computation such that the computation is in fact highly
modularised \citep{bushnaq2024local_interaction_basis}.
However, efforts to find such modularity have proven difficult \citep{filan2021clusterability},
partially because it is not clear in what form we should expect such modularity to appear.

One way to understand the modularity that neural networks naturally and implicitly form,
is to enforce some modular computation and see how the neural networks react.
Sparse MoE models have this kind of enforced modularity and we hope that in studying
these models we can develop better tools for uncovering possible latent modularity in dense neural networks.
In particular, one promising path for understanding modularity may be through the optimisation pressures
that resulted in such modularity.
For example specialisation and reducing connection costs are two evolutionary pressures for
modularity to develop in biological models \citep{clune2013evolution_of_modularity, mcdougall2022modularity}.
Analogously, we might expect that specialisation pressures in the training processes for AI systems
also results in the development of modularity as in \citet{wang2025differentiation_and_specialisation_of_attention}.

\section{SAE Features as Routing Predictors}
\label{sec:sae_routing_interface}
We demonstrate that SAE features can effectively predict routing patterns.

\paragraph{Routing Probes.}
We extract activations $\vx$ from the residual stream before each layer's attention block and
use an SAE encoder to produce sparse feature representations $\vz$.
We then train a logistic regression classifier (one per expert $E_i$)
on these sparse feature representations using 1M tokens total
to predict whether a given expert is activated for some context.
Each classifier is trained with binary cross-entropy against the ground-truth binary label
indicating whether $E_i$ appears in the router's top-$k$ selection for that token.
We report \emph{macro-F1}: the mean per-expert F1 for this binary classification.\footnote{
  We additionally show that classifiers recover the router's full probability ordering,
  not just top-$k$ membership in \Cref{sec:appendix:routing_correlation}.
}
Routing-prediction probes that restrict how many SAE features are active per token
are in \Cref{sec:appendix:num_features_ablation}.

\paragraph{Baselines.}
Previous work interpreting MoE routing like \citet{jiang2024mixtral} and \citet{zoph2022st_moe} used a token-level analysis that we denote the
\textbf{unigram baseline}.
For each token in the vocabulary,
we count how often it co-occurs with each expert's activation on a training set,
and at inference predict the experts it most frequently co-occurred with.
We extend prior unigram-only analyses to also include \textbf{bigrams} (current and previous token),
allowing us to assess whether local token context improves prediction.
To distinguish whether prediction gains stem from the learned SAE basis
or simply from access to continuous residual stream representations,
we additionally compare against sparse linear probes on two alternative bases.
The \textbf{neuron basis probe} retains only the top-$s$ highest-activating neurons
from the residual stream, setting all others to zero,
and trains a linear classifier on this sparse representation.
The \textbf{PCA basis probe} first projects the residual stream
onto its leading principal components,
then applies the same top-$s$ sparsification in this rotated basis.
Here probe sparsity is required to fairly compare against the sparse SAE-based classifier;
we match $s{=}32$ on OLMoE and $s{=}128$ on gpt-oss.

\paragraph{SAE Latents Predict Expert Routing.}
\label{sec:results:routing}

\begin{table}[t!]
  \centering
  \begin{tabular}{lcc}
    \toprule
    & \textbf{OLMoE-1B-7B} & \textbf{gpt-oss-20b} \\
    \midrule
    Unigram Baseline & 0.564 & 0.296 \\
    Bigram Baseline & 0.633 & 0.356 \\
    Neuron Basis Probe & 0.666 & 0.586 \\
    PCA Basis Probe & 0.680 & 0.536 \\
    SAE Predictor (Ours) & \textbf{0.740} & \textbf{0.730} \\
    \bottomrule
  \end{tabular}
  \caption{SAE latents predict MoE routing better than token-level statistics and sparse probes.
    Neuron and PCA basis probes operate on sparse residual stream representations
    matched in sparsity to the SAE.
    N-gram baselines predict routing from token co-occurrence frequencies.
  Results reported over $\sim$1M tokens as macro-F1 for layer 11 of OLMoE-1B-7B and layer 12 of gpt-oss-20b.}
  \label{tab:routing-prediction-both}
\end{table}

To show that learned sparse features are capable of predicting expert routing accurately,
we compare SAE-based prediction against n-gram baselines and sparse probes.
\Cref{tab:routing-prediction-both} shows two findings.
First, token-level statistics are a poor proxy for expert routing,
with unigram and bigram baselines lagging far behind SAE latents on both OLMoE-1B-7B and gpt-oss-20b.
Second, at matched sparsity the SAE predictor also outperforms both neuron-basis and PCA-basis sparse probes on each model,
showing the gain is specific to the learned SAE basis rather than any linear readout of the residual stream.
We show in \Cref{sec:appendix:routing_prediction_layers} that this SAE-basis advantage
extends across multiple layers and sparsity levels ($s{\in}\{64,128\}$) of gpt-oss-20b,
where the SAE beats the best neuron and PCA probe at four of five layers.
Since all three probes use the same residual activations and differ only in the basis where sparsity is enforced,
this supports the view that the SAE features carry routing-specific structure, motivating RouterInterp's use of SAE decomposition.

\section{Feature Selection Method and Set Size}
\label{sec:appendix:feature_selection}
\subsection{Feature Selection Method}
\label{sec:appendix:feature_selection:method}

To identify which SAE features support each expert's activation,
we use gradient-based attribution
to estimate each active feature's effect on routing to that expert.
As in \Cref{sec:background:moe}, the router produces logits $h(\vx)$
and selects a set $\mathcal{T}(\vx)$ of $k$ experts.
For expert $E_i$, let $\tau_i(\vx)$ be the $k$-th largest logit among all \emph{other} experts.
The value $\tau_i(\vx)$ is the selection threshold that $E_i$ must exceed.
The gap between $E_i$'s logit and this threshold tells us how close the expert is to entering the top-$k$
and gives us a continuous quantity to attribute to SAE features.
We define this gap as the routing margin:
\begin{align}
  \label{eq:routing-margin}
  m_i(\vx) = h(\vx)_i - \tau_i(\vx),
\end{align}
Following SAE feature-attribution methods~\citep{templeton2024scaling_monosemanticity,marks2025sparse_feature_circuits,shu-etal-2025-beyond}
we approximate the effect of ablating feature $f$ on whether $E_i$ is selected by
\begin{align}
  \label{eq:routing-margin-attr}
  c_{if}(\vx) = z_f\, \vd_f^\top \nabla_{\vx} m_i(\vx),
\end{align}
where $z_f = (\vz)_f$ is the feature activation and $\vd_f$ is the corresponding column of $\Wdec$.
Thus, ablating the feature changes the margin by approximately $-c_{if}$,
so positive values of $c_{if}$ indicate that the feature supports selecting $E_i$.
We favour features whose positive contribution is specific to tokens routed to $E_i$
and discount contributions that also occur when $E_i$ is not selected:
\begin{align}
  \label{eq:attribution-score}
  S_{if} = \mathbb{E}[c_{if}^{+}\mid E_i\in\mathcal{T}(\vx)]
  - \mathbb{E}[c_{if}^{+}\mid E_i\notin\mathcal{T}(\vx)],
\end{align}
where $c_{if}^{+}=\max(c_{if},0)$\footnote{
  This score does not account for SAE reconstruction error,
  or the possibility that ablating a feature changes which expert sets the selection threshold $\tau_i(\vx)$.
}.
For each expert $E_i$,
we construct a feature set $\sF_i$
that has no overlap with any other expert's feature set.
This isolates the routing signals that distinguish one expert from another.
Each feature is assigned to at most one expert, preferring the highest attribution score $S_{if}$,
and we keep the top-$n$ assigned features per expert.

\subsection{Selection Criterion and Set Size}
\label{sec:appendix:feature_selection:criterion}

Above we select the top-$n$ features by the attribution score $S_{if}$ (\Cref{eq:attribution-score}) for each expert.
Here we compare this choice to two simpler alternatives,
and evaluate how the feature set size $n$ affects routing-prediction probe F1.

The first alternative is $\rho$-usefulness \citep{ilyas2019rho_usefulness},
a data-driven criterion that measures how discriminative a feature is for expert selection.
A feature $f$ is $\rho$-useful for expert $E_i$ if its expected activation is higher
when $E_i$ is selected than when it is not:
\begin{align*}
  \rho(f,E_i)=\mathbb{E}[f(\vx)\mid E_i\in\mathcal{T}(\vx)]-\mathbb{E}[f(\vx)\mid E_i\notin\mathcal{T}(\vx)].
\end{align*}
Features with high $\rho$-usefulness are specifically predictive of that expert's activation.
The attribution score $S_{if}$ applies the same contrast to margin attributions $c_{if}^{+}$,
rather than to raw activations.

The second alternative ranks features by geometric alignment
between SAE decoder directions $\vd_f$ (columns of $\Wdec$) and router weight vectors $\vw_i$ (rows of $\mW_r$).
For each expert $E_i$, we rank features $f$ by $\cos(\vd_f, \vw_i)$ and select the top-$n$.

To compare the three criteria, we train a linear probe on a dataset-level set of $n$ features:
the same $n$ features are used on every token, with all others zeroed.
We train only on tokens where at least one of those $n$ features is active.
This serves as a proxy for RouterInterp explanations,
where the explainer LLM only sees examples from a fixed set of features.

\begin{figure}[H]
  \centering
  \includegraphics[width=0.9\linewidth]{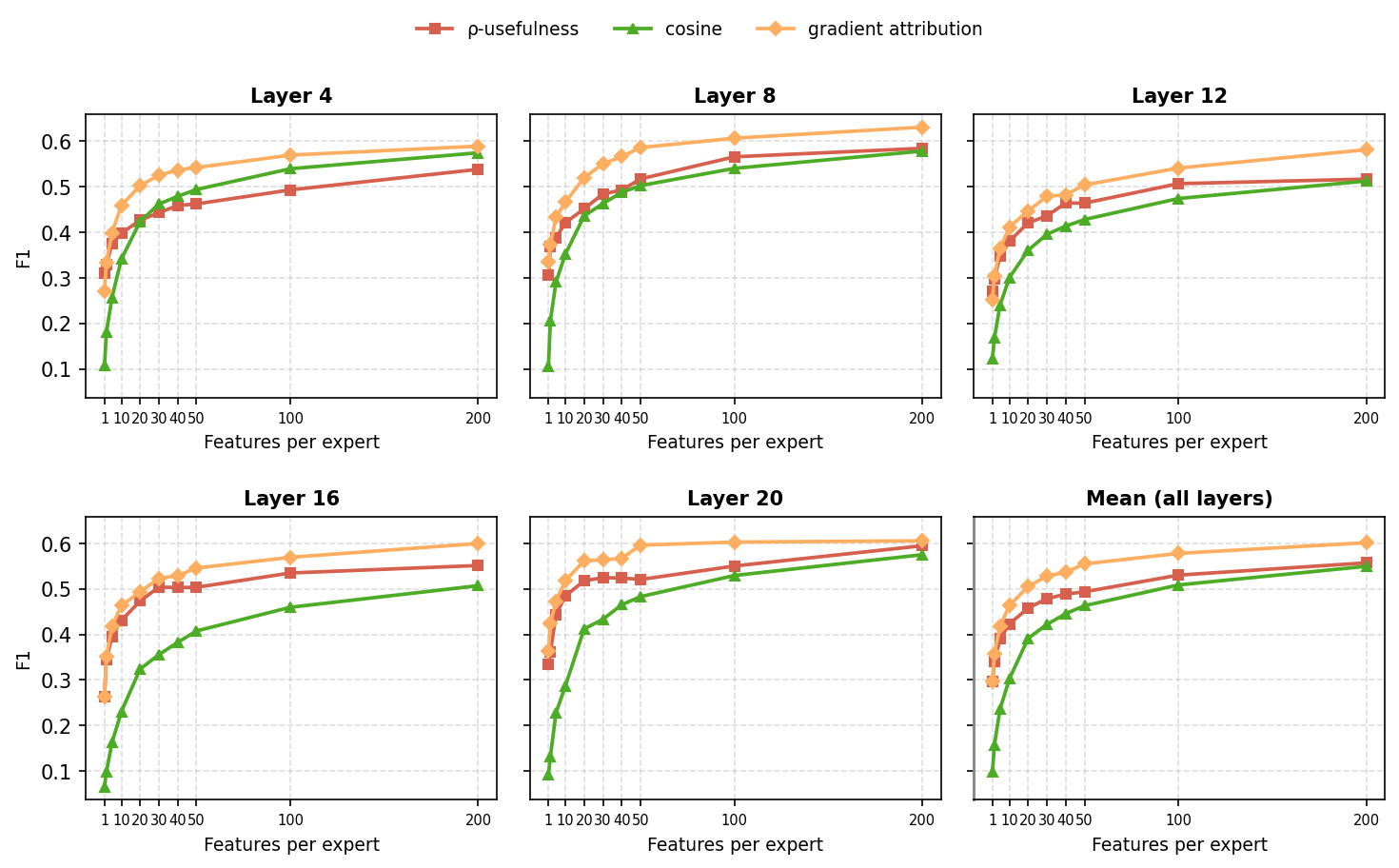}
  \caption{
    Gradient-based attribution is the strongest feature-selection criterion
    across five layers of gpt-oss-20b and every feature set size $n$.
    For each feature selection method,
    we select the top-$n$ features per expert
    and train a linear probe to predict routing to an expert
    using only these features' activations.
    The mean F1 across layers improves only gradually beyond ${\sim}50$ features.
  }
  \label{fig:feature_selection_sweep}
\end{figure}

\Cref{fig:feature_selection_sweep} shows that gradient-based attribution
outperforms both alternatives across all five layers of gpt-oss-20b.
$\rho$-usefulness comes closest,
while cosine similarity lags behind at small $n$ before all three converge at large $n$.
The gradual improvement of the mean beyond ${\sim}50$ features
indicates that routing-relevant information is redundantly encoded across many SAE features.

\begin{figure}[h]
  \centering
  \includegraphics[width=0.8\linewidth]{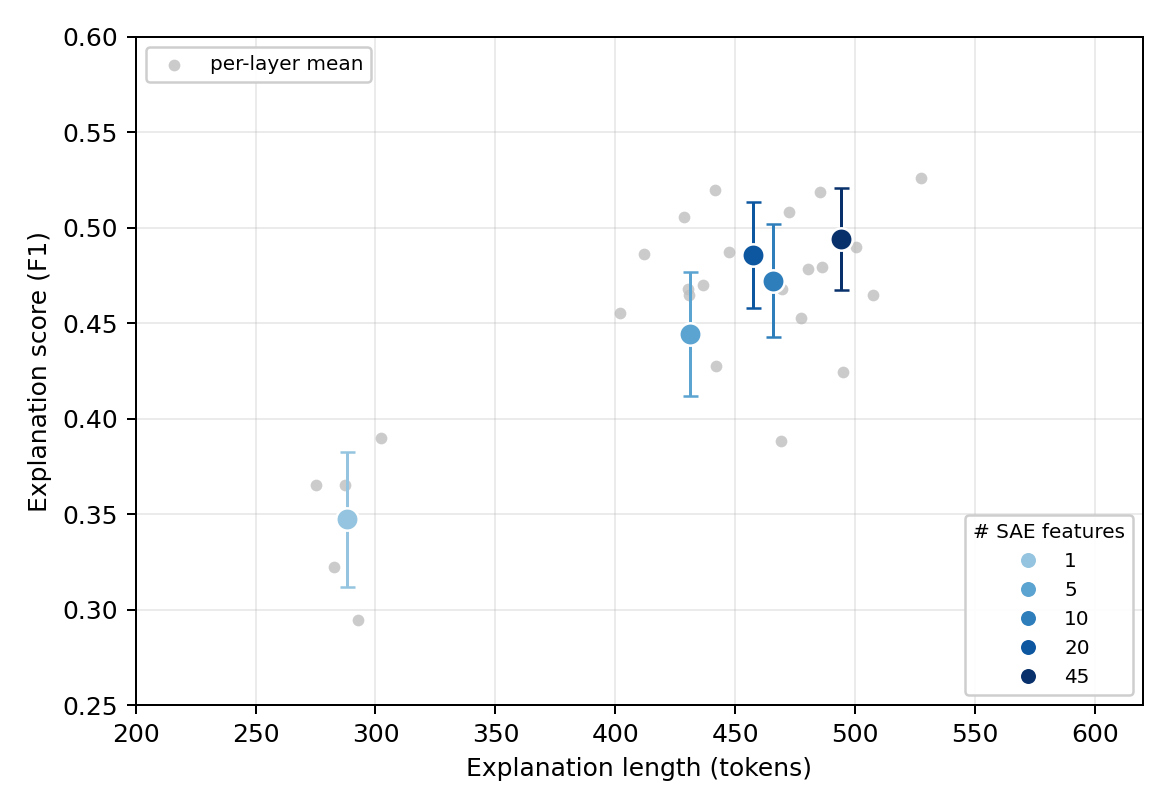}
  \caption{
    Up to $n{=}45$ features,
    every additional feature still improves RouterInterp explanations,
    but the gains shrink as descriptions grow.
    Explanation length grows roughly linearly with the feature set size $n$,
    while the explanation score improves steeply at small $n$ and then slows:
    from $n{=}25$ to $n{=}45$, descriptions grow by ${\sim}90$ tokens while the score improves by only ${\sim}0.03$.
    Each point is a feature set size $n \in \{1, 5, 15, 25, 45\}$,
    with score and length averaged across experts over 5 layers of gpt-oss-20b.
  }
  \label{fig:f1_vs_tokens}
\end{figure}

For RouterInterp explanations, we select $n{=}45$ features per expert.
This is a deliberate tradeoff between predictive coverage and explainability:
as \Cref{fig:f1_vs_tokens} shows, explanation length grows roughly linearly with $n$
while explanation score gains diminish.
Extrapolating this linear trend,
including substantially more features (e.g.\ $n{=}200$) would produce descriptions
several times longer and proportionally more expensive to generate,
reducing interpretability for human readers without a comparable gain in predictive performance
\citep{sharkey2024sparsify, ayonrinde2024mdl_saes, ayonrinde2025evaluatingexplanations}.

\section{Expert Routing Across Pile Domains}
\label{sec:appendix:pile_domains}
\begin{figure}[t]
  \centering
    \includegraphics[width=\linewidth]{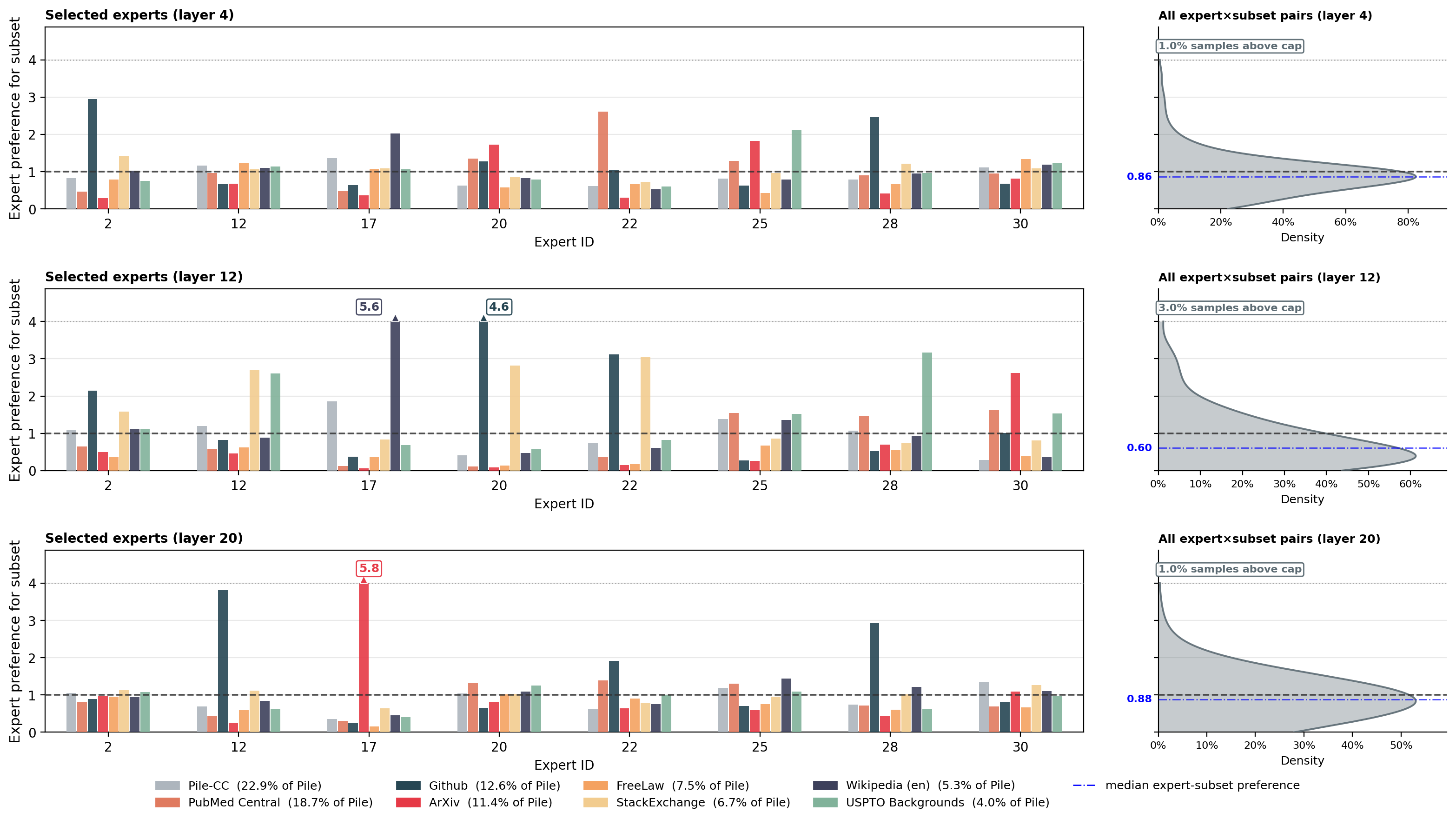}
    \caption{
        \textbf{Expert routing distributions across Pile domains provide
        evidence against macro-domain DSH and are consistent with SSH.}
        Adapting the Mixtral analysis of \citet{jiang2024mixtral} on gpt-oss-20b,
        we plot how often an expert's routed tokens come from a subset,
        relative to this subset's fraction of the corpus (the
        \emph{expert routing preference}).
        We show routing preferences for randomly selected experts
        across 16 subsets (left) and their density over all
        expert--subset pairs (right), at an early (layer~4), a middle
        (layer~12), and a late (layer~20) layer.
        The densities are tightly concentrated at the corpus rate (dashed line at~$1$).
        If expert domains coincided with Pile subsets,
        preferences would instead spread far from~$1$:
        near zero for most pairs,
        and at many times the corpus rate for a few dedicated specialists.
        Values above 4.0 are capped for better readability.
    }
\label{fig:pile_domain}
\end{figure}

The DSH and SSH make competing claims about how experts specialise over
\emph{domains}: under the DSH, an expert concentrates on a coherent domain;
under the SSH, it spans a disjoint collection of micro-domains.
These domains may be semantic, syntactic, or otherwise defined by the router,
and are not given a priori.
Some corpora, however, provide labelled subsets that can serve as a coarse
proxy for domain boundaries---for example, the Pile's
\citep{gao2020pile} partitions into ArXiv, Github, PubMed, and related sources.

\citet{jiang2024mixtral} explored whether Mixtral experts specialise in these Pile subsets,
measuring, for each expert, what fraction of its routed tokens come from
each subset.
They found little systematic variation in token-to-expert assignment across subsets
and concluded that routing follows syntax more closely than topic.
If the DSH holds and its domains coincide with Pile subsets, one would
expect clear subset specialisation: tokens from a given subset should be
disproportionately routed to particular experts.
Mixtral's finding of little variation is therefore already evidence against
that form of the DSH.

We replicate their analysis on a newer MoE model.
We process ${\sim}$10M tokens from 16~Pile subsets spanning
scientific, legal, code, and web data.
For each expert--subset pair we compare how often the expert's routed
tokens come from the subset with how often the subset's tokens occur
in the corpus overall.
We compute the \emph{expert routing preference} for the subset, $p(u\,|\,E_i)\,/\,p(u)$:
the fraction of expert $E_i$'s routed
tokens that come from subset~$u$,
over the subset's fraction of all tokens.
A value of~$1$ means the expert receives the subset's tokens exactly at the corpus rate.
\Cref{fig:pile_domain} shows results for
early, middle, and late layers of gpt-oss-20b.
The left panels show randomly selected experts from each layer, and
some exhibit strong spikes on particular subsets (e.g.\ Github or ArXiv).
Individual experts are only illustrative, however,
and the layer-wide density shows the real picture.
Routing preferences show little spread around the corpus rate:
for most expert--subset pairs in a layer,
a subset's tokens reach the expert at or below the corpus rate.
This pattern is consistent with the SSH.
If each expert's fine-grained micro-domains are distributed roughly proportionally across Pile domains,
routing at the subset level should remain close to corpus-proportional for most pairs.
Under the DSH, by contrast,
each subset's tokens would concentrate on a few dedicated experts,
producing a bimodal distribution:
most expert--subset pairs would sit near~$0$
(non-specialist experts receive almost none of the subset's tokens),
while a few pairs would lie at many times the corpus rate (the dedicated specialists).

The observed distribution is therefore further evidence against that
form of the DSH in which expert domains align with the Pile subsets.
The relevant domains could cut across the Pile's dataset boundaries or be
defined by properties such as syntax rather than topic,
in which case one would not expect tokens from particular Pile subsets
to be preferentially routed to particular experts.
Nevertheless, for many plausible semantic domain decompositions,
we would expect at least some Pile subsets to correlate with the domain boundaries.

\section{Routing Prediction Across Layers}
\label{sec:appendix:routing_prediction_layers}
We extend the matched-sparsity probe comparison of \Cref{sec:results:routing}
to five layers of gpt-oss-20b at two sparsity levels, $s{\in}\{64, 128\}$.
\Cref{tab:routing-prediction-oss} reports macro-F1 for SAE, neuron, and PCA sparse linear probes.
At $s{=}128$, the SAE predictor outperforms the best neuron and PCA probe at four of five layers,
showing that the SAE-basis advantage holds across both depth and sparsity.
We compare two SAE variants ($s{=}64$ and $s{=}128$);
the effect of the underlying SAE is discussed in \Cref{sec:appendix:sae_metrics}.

\begin{table}[h]
    \centering
    \begin{tabular}{lccccc}
    \toprule
     & \textbf{Layer 4} & \textbf{Layer 8} & \textbf{Layer 12} & \textbf{Layer 16} & \textbf{Layer 20} \\
    \midrule
    SAE Predictor ($s{=}128$) & \textbf{0.569} & 0.702 & \textbf{0.730} & 0.439 & \textbf{0.789} \\
    SAE Predictor ($s{=}64$) & 0.552 & \textbf{0.725} & 0.685 & 0.297 & 0.773 \\
    Neuron Probe ($s{=}128$) & 0.519 & 0.638 & 0.586 & \textbf{0.559} & 0.630 \\
    Neuron Probe ($s{=}64$) & 0.476 & 0.603 & 0.509 & 0.524 & 0.570 \\
    PCA Probe ($s{=}128$) & 0.544 & 0.623 & 0.536 & 0.494 & 0.366 \\
    PCA Probe ($s{=}64$) & 0.517 & 0.603 & 0.523 & 0.482 & 0.384 \\
    \bottomrule
    \end{tabular}
    \caption{The learned SAE basis captures more routing-relevant structure
    than standard bases at matched sparsity across most layers of gpt-oss-20b.
    All methods use sparse linear probes with matched sparsity ($s$ active dimensions).
    At $s{=}128$, the SAE predictor achieves strictly higher macro-F1 than the best neuron or PCA probe
    at four of five layers.
    All values reported as macro-F1.}
    \label{tab:routing-prediction-oss}
\end{table}

\section{Routing Depends on Multiple Features}
\label{sec:appendix:num_features_ablation}
Routing prediction in \Cref{sec:results:routing} uses the full set of active SAE features per token.
Here we show that this routing information is spread across many latents,
by ablating how many of those features are kept.
Let $m$ be the number of active SAE features kept per token
(e.g.\ $m{=}32$ means 32 of the $s{=}128$ latents on that token are kept, and the rest are zeroed).
The routing-prediction table in \Cref{sec:sae_routing_interface} corresponds to the full-sparsity probe
($m{=}s{=}128$ on gpt-oss), i.e.\ $m{=}128$ here.
Unlike \Cref{sec:appendix:feature_selection:criterion}, which varies a dataset-level selected set of $n$ features,
this ablation varies the per-token active count.
Table~\ref{tab:routing-prediction-m} shows macro-F1 for varying $m \in \{1, 8, 16, 32, 64, 128\}$ on gpt-oss-20b with SAE sparsity $s=128$.

\begin{table}[!h]
  \centering
  \begin{tabular}{lccccc}
    \toprule
    & \textbf{Layer 4} & \textbf{Layer 8} & \textbf{Layer 12} & \textbf{Layer 16} & \textbf{Layer 20} \\
    \midrule
    Unigram Baseline & 0.472 & 0.442 & 0.296 & 0.343 & 0.369 \\
    Bigram Baseline & 0.535 & 0.511 & 0.356 & 0.406 & 0.411 \\
    \midrule
    SAE Predictor ($m=1$) & 0.183 & 0.260 & 0.185 & 0.089 & 0.280 \\
    SAE Predictor ($m=8$) & 0.235 & 0.380 & 0.267 & 0.249 & 0.617 \\
    SAE Predictor ($m=16$) & 0.349 & 0.408 & 0.434 & 0.387 & 0.694 \\
    SAE Predictor ($m=32$) & 0.490 & 0.640 & 0.632 & 0.414 & 0.749 \\
    SAE Predictor ($m=64$) & \textbf{0.576} & 0.699 & 0.694 & 0.400 & 0.777 \\
    SAE Predictor ($m=128$) & 0.569 & \textbf{0.702} & \textbf{0.730} & \textbf{0.439} & \textbf{0.789} \\
    \bottomrule
  \end{tabular}
  \caption{
    Routing prediction improves significantly as more SAE features are used,
    consistent with polysemantic expert behaviour predicted by the SSH.
    Results are shown for gpt-oss-20b with SAE sparsity $s=128$.
    Performance improves with $m$ across all reported layers, peaking at $m{=}128$
    in four of five layers,
    indicating that routing depends on multiple distinct concepts rather than a single domain.
  }
  \label{tab:routing-prediction-m}
\end{table}

The results reveal a clear pattern:
macro-F1 generally improves as $m$ increases across layers.
Layer 16 rises from 0.089 ($m=1$) to 0.439 ($m=128$),
and Layer 20 rises from 0.280 ($m=1$) to 0.789 ($m=128$).
This trend indicates that routing decisions depend on multiple distinct concepts,
consistent with polysemantic expert behaviour.

\section{SAE Setup and Reconstruction Quality}
\label{sec:appendix:sae_metrics}
In this section, we report SAE setup and reconstruction quality.
SAEs are trained on residual-stream activations before each layer's attention block.

\paragraph{OLMoE-1B-7B.}
We train Top-K SAEs using the Sparsify library
on activations from layers 3, 7, 11, and 15 of OLMoE-1B-7B,
sampled from 100M tokens of OLMoE-mix-0924~\citep{muennighoff2025olmoe}.
Each SAE has 32,768 features (16$\times$ expansion) with per-token latent sparsity $s{=}32$.
Training all four SAEs took approximately 2 hours in total on a single A100 GPU.
\Cref{tab:sae-metrics-olmoe} reports the fraction of variance unexplained (FVU)
and the percentage of dead features (features that never activate on the evaluation set).

\begin{table}[h]
    \centering
    \begin{tabular}{lcc}
    \toprule
    \textbf{Layer} & \textbf{FVU} & \textbf{Dead features (\%)} \\
    \midrule
    3  & 0.007 & 4.11 \\
    7  & 0.024 & 0.51 \\
    11 & 0.064 & 0.59 \\
    15 & 0.218 & 0.51 \\
    \bottomrule
    \end{tabular}
    \caption{SAE reconstruction quality for OLMoE-1B-7B (Top-K; 32,768 features, 16$\times$ expansion, $s{=}32$).
    FVU increases with depth while dead feature rates remain low.}
    \label{tab:sae-metrics-olmoe}
\end{table}

\paragraph{gpt-oss-20b.}
For gpt-oss-20b, we use publicly available BatchTopK SAEs
from \citet{lin2025neuronpedia_fall},\footnote{\url{https://huggingface.co/andyrdt/saes-gpt-oss-20b}}
trained on layers 4, 8, 12, 16, and 20.
Each SAE has 131,072 features (${\sim}$45$\times$ expansion)
with two per-token latent sparsity variants: $s{=}64$ and $s{=}128$.
\Cref{tab:sae-metrics-oss} reports FVU and dead feature rates for both variants.

\begin{table}[h]
    \centering
    \begin{tabular}{lcccc}
    \toprule
    & \multicolumn{2}{c}{\textbf{FVU}} & \multicolumn{2}{c}{\textbf{Dead features (\%)}} \\
    \cmidrule(lr){2-3} \cmidrule(lr){4-5}
    \textbf{Layer} & $s{=}64$ & $s{=}128$ & $s{=}64$ & $s{=}128$ \\
    \midrule
    4  & 0.050 & 0.037 & 0.04 & 0.90 \\
    8  & 0.101 & 0.077 & 0.02 & 0.29 \\
    12 & 0.172 & 0.137 & 0.00 & 0.01 \\
    16 & 0.232 & 0.191 & 0.00 & 0.00 \\
    20 & 0.146 & 0.117 & 0.02 & 0.00 \\
    \bottomrule
    \end{tabular}
    \caption{SAE reconstruction quality for gpt-oss-20b (BatchTopK; 131,072 features, ${\sim}$45$\times$ expansion).
    FVU peaks at layer~16 for both sparsity levels;
    $s{=}128$ consistently achieves lower FVU than $s{=}64$.}
    \label{tab:sae-metrics-oss}
\end{table}

\paragraph{RouterInterp quality is downstream of SAE quality.}
SAE reconstruction quality correlates with routing prediction performance across layers.
This is most visible at layer~16 of gpt-oss-20b,
where FVU is highest (0.232 for $s{=}64$, 0.191 for $s{=}128$)
and routing-prediction macro-F1 is weakest (0.297 for $s{=}64$; 0.439 for $s{=}128$).
Explanation scores do not show the same dependence:
they pass through LLM generation and scoring, which can mask the effect of SAE reconstruction quality.
Improving SAE quality at this layer---FVU drops from 0.232 to 0.191
when moving from $s{=}64$ to $s{=}128$---yields the largest gain
in routing prediction (0.297 $\to$ 0.439).
The co-occurrence of high FVU and weak routing prediction suggests that poor reconstruction may be a bottleneck:
probe performance can depend on SAE quality,
and better SAEs would improve RouterInterp's performance.

\section{Human Evaluation}
\label{sec:appendix:human_interpretability}
In \Cref{sec:methods:scoring}, we introduced an LLM-based scorer to evaluate RouterInterp explanation quality as a proxy for human judgment.
To validate whether this LLM scorer faithfully reflects human judgment,
we conducted a blind human evaluation
in which human raters performed the same binary labelling task as the LLM scorer,
allowing us to measure how consistently human raters agree with each other (inter-annotator agreement)
and whether the LLM scorer matches human judgments.
We collected annotations on 150 text/context pairs (6 examples per expert across 25 experts)
from layer 12 of gpt-oss-20b.
Each example was labelled by two independent human evaluators and by the LLM scorer.
We report pooled agreement and Cohen's $\kappa$ with uncertainty estimated over repeated resampling
(\Cref{tab:human_eval_agreement}).

\begin{table}[h]
  \centering
  \begin{tabular}{lcc}
    \toprule
    Metric & Human--Human & Human--LLM \\
    \midrule
    Pooled agreement & 0.807 $\pm$ 0.070 & 0.857 $\pm$ 0.050 \\
    Cohen's $\kappa$ & 0.487 $\pm$ 0.165 & 0.602 $\pm$ 0.150 \\
    \bottomrule
  \end{tabular}
  \caption{The LLM scorer is at least as consistent a judge as a human rater.
    Pooled agreement is the fraction of examples where both raters gave the same label;
    Cohen's $\kappa$ is a stricter metric that accounts for the fact that
    two raters could coincidentally agree just by both labelling most examples the same way.
    Values are mean $\pm$ std over repeated resampling;
  measured on 150 examples across 25 experts from layer~12 of gpt-oss-20b.}
  \label{tab:human_eval_agreement}
\end{table}

Human--LLM agreement is comparable to Human--Human agreement on both metrics,
with no statistically significant difference.
Because only two human annotators were compared per example,
some disagreement is expected from subjective thresholds on edge cases.
Despite this limited study scale,
the results support using our LLM scorer as a practical and
reliable proxy for human interpretability judgments.

Our findings are also consistent with prior evidence that automated interpretation and
LLM-as-a-judge methods correlate with human raters and can scale beyond manual-only pipelines
\citep{paulo2025automatically, zheng2023judging_llm_as_a_judge, anthis2025llm_social_simulations}.

\section{SAE Probes Recover Router Probabilities}
\label{sec:appendix:routing_correlation}
In \Cref{sec:results:routing}, we evaluated SAE-based routing prediction using Macro-F1.
As an additional validation of the SAE predictor's capabilities,
we check whether the predicted probabilities also align with the router's continuous output.
The classifiers are trained only with binary cross-entropy on top-$k$ membership
and are not directly optimised to recover the full probability distribution.
For each token, we take the vector of predicted probabilities
and the vector of true router probabilities (one entry per expert)
and compute their Spearman rank correlation and MSE.

\begin{table}[t]
    \centering
    \small
    \begin{tabular}{lccc}
    \toprule
    \textbf{Layer} & \textbf{Macro-F1} & \textbf{Spearman $\rho$} & \textbf{MSE} \\
    \midrule
    4  & 0.569 & 0.258 & 0.0278 \\
    8  & 0.702 & 0.809 & 0.0294 \\
    12 & 0.730 & 0.741 & 0.0295 \\
    16 & 0.439 & 0.413 & 0.0285 \\
    20 & 0.789 & 0.817 & 0.0296 \\
    \bottomrule
    \end{tabular}
    \caption{%
    The predicted probabilities from SAE-based probes align well with the actual MoE router probabilities,
    even though the probes are trained only with binary cross-entropy on top-$k$ membership.
    Spearman $\rho$ is the rank correlation between the predicted and true probability assigned to each expert,
    computed per token and averaged; high values mean the probe recovers which experts the router assigns more probability to.
    MSE is the mean squared error between predicted and true per-expert probabilities ($\approx 0.03$; RMSE $\approx 0.17$).
    Spearman correlates with macro-F1 across layers (both strong at 8, 12, 20; weaker at 4 and 16).
    gpt-oss-20b; SAE predictor with $s{=}128$.}
    \label{tab:probe-router-correlation}
\end{table}

\Cref{tab:probe-router-correlation} shows that at layers where classification is strong
(8, 12, 20), Spearman $\rho$ exceeds 0.7,
indicating that the probe recovers not just which experts are selected
but their relative ranking.
At layers 4 and 16, where macro-F1 is weaker,
calibration drops accordingly.
MSE is uniformly low (${\approx}\,0.03$; RMSE ${\approx}\,0.17$),
showing that predicted probabilities stay close to the true router values
even at weaker layers.

\section{Prompts}
\label{sec:appendix:prompts}
We provide the prompts used for expert-level explanations,
the n-gram explanation baseline,
and automatic explanation scoring (\Cref{sec:methods:routerinterp}).

\paragraph{RouterInterp.}
\label{app:prompts:aggregation}
RouterInterp prompts the explainer on context windows grouped by feature:
for each selected feature, positives where the feature fires and the expert is selected,
and hard negatives where the feature fires but the expert is not selected.

\begin{promptlisting}
You are a meticulous AI researcher conducting an important investigation into patterns found in language. Your task is to analyze text and provide an explanation that thoroughly encapsulates possible patterns found in it.

You are given text examples with the relevant token(s) wrapped in <<double angle brackets>> (a run of consecutive tokens may be wrapped together). These are contexts where the expert was active. When routing weights are shown in parentheses after an example, they indicate how strongly each token was activated. If the examples are organised into numbered groups, each group was collected from a different SAE feature --- but multiple groups may share the same underlying activation pattern.

Each group comes from one SAE feature associated with this expert. Routed examples show windows where that feature fires AND this expert is selected. Hard-negative examples (under a "Hard negatives (feature fires, expert not selected):" header) show windows where the SAME feature fires but this expert is NOT selected. Use the contrast to describe the routing-relevant context --- what makes the expert fire given the feature --- not merely where the feature activates. Write a positive, specific description; do not frame the output as "activates on X but not Y".

HOW TO WRITE A GOOD DESCRIPTION:
- MERGE SIMILAR GROUPS. The groups are raw feature data, not pre-identified modes. Look across all groups first: if several groups show the same pattern (same token type, same syntactic role, same domain), merge them into a single description. Only describe as many modes as there are genuinely distinct patterns in the data --- this may be fewer than the number of groups, or even just one.
- FIND PATTERNS, DON'T JUST LIST TOKENS. The highlighted tokens are clues, not the answer. Generalise: state the rule they share --- their part of speech, grammatical or structural role, the domain or topic, the format, or what comes before/after them. Concrete examples are welcome when they help a human grasp the pattern (quote a couple of representative tokens or phrases), but a description that is ONLY a list of tokens is useless.
- DESCRIBE THE CONTEXT, NOT THE SURFACE TOKEN. The same surface token appears in both activating and non-activating windows, so token identity alone is never enough --- what matters is the precise context that makes the expert fire. For example, write ``"(" immediately following a function name in Python code'', not ``the token "("''.
- LET LENGTH FOLLOW THE DATA. A single clean pattern needs one sentence. A genuinely polysemantic expert needs one clear description per truly distinct mode. Do NOT force unrelated modes into one summary, and do NOT drop a real mode just to be brief. Avoid repetition and padding.
- Avoid announcing a count of modes (e.g. "activating in ten distinct contexts") --- just describe the modes you actually find.
- Avoid generic catch-alls like "common function words", "technical text", or "structural delimiters" --- those describe nothing and match everything.
- Do not use numbered lists or bullet points, and do not mention the <<>> markers.
- NEVER reuse token patterns, surface forms, or domains from the few-shot demonstration --- describe only what appears in the actual examples provided.
- If the examples share no specific distinguishing pattern, say so plainly (e.g. "diverse general English prose with no single trigger"); an honest broad description is correct, an invented specific pattern is a failure.

Respond directly with the explanation.
\end{promptlisting}

\paragraph{Expert Activations AutoInterp.}
\label{app:prompts:expert-autointerp}
The same base prompt, with a non-routed grouped contrastive section:
non-activating windows under a ``Non-activating examples:'' header rather than routed hard negatives.

\begin{promptlisting}
You are a meticulous AI researcher conducting an important investigation into patterns found in language. Your task is to analyze text and provide an explanation that thoroughly encapsulates possible patterns found in it.

You are given text examples with the relevant token(s) wrapped in <<double angle brackets>> (a run of consecutive tokens may be wrapped together). These are contexts where the expert was active. When routing weights are shown in parentheses after an example, they indicate how strongly each token was activated. If the examples are organised into numbered groups, each group was collected from a different SAE feature --- but multiple groups may share the same underlying activation pattern.

Non-activating examples (same highlighted token, different context where the expert did NOT fire) are shown after each group's activating examples under a "Non-activating examples:" header. Use them to sharpen your description of the activating pattern --- they reveal which aspects of the context are incidental and which are essential. Write a positive, specific description of what the expert activates on; do not frame the output as "activates on X but not Y". Instead, state precisely what makes the activating context specific (e.g. "specifically at function call sites" rather than "in code but not in prose").

HOW TO WRITE A GOOD DESCRIPTION:
- MERGE SIMILAR GROUPS. The groups are raw feature data, not pre-identified modes. Look across all groups first: if several groups show the same pattern (same token type, same syntactic role, same domain), merge them into a single description. Only describe as many modes as there are genuinely distinct patterns in the data --- this may be fewer than the number of groups, or even just one.
- FIND PATTERNS, DON'T JUST LIST TOKENS. The highlighted tokens are clues, not the answer. Generalise: state the rule they share --- their part of speech, grammatical or structural role, the domain or topic, the format, or what comes before/after them. Concrete examples are welcome when they help a human grasp the pattern (quote a couple of representative tokens or phrases), but a description that is ONLY a list of tokens is useless.
- DESCRIBE THE CONTEXT, NOT THE SURFACE TOKEN. The same surface token appears in both activating and non-activating windows, so token identity alone is never enough --- what matters is the precise context that makes the expert fire. For example, write ``"(" immediately following a function name in Python code'', not ``the token "("''.
- LET LENGTH FOLLOW THE DATA. A single clean pattern needs one sentence. A genuinely polysemantic expert needs one clear description per truly distinct mode. Do NOT force unrelated modes into one summary, and do NOT drop a real mode just to be brief. Avoid repetition and padding.
- Avoid announcing a count of modes (e.g. "activating in ten distinct contexts") --- just describe the modes you actually find.
- Avoid generic catch-alls like "common function words", "technical text", or "structural delimiters" --- those describe nothing and match everything.
- Do not use numbered lists or bullet points, and do not mention the <<>> markers.
- NEVER reuse token patterns, surface forms, or domains from the few-shot demonstration --- describe only what appears in the actual examples provided.
- If the examples share no specific distinguishing pattern, say so plainly (e.g. "diverse general English prose with no single trigger"); an honest broad description is correct, an invented specific pattern is a failure.

Respond directly with the explanation.
\end{promptlisting}

\paragraph{Expert Impact AutoInterp.}
\label{app:prompts:expert-impact}
We use exactly the same prompt as \citet{herbst2026expert} and do not reproduce it here.

\paragraph{N-gram AutoInterp.}
\label{app:prompts:ngram}
For Unigram AutoInterp and Bigram AutoInterp, the model compresses frequent co-occurring tokens into a natural-language description.

\begin{promptlisting}
    Your job is to look for patterns in text. You will be given a list of WORDS, your task is to provide an explanation for what pattern best describes them. Here are some guidelines:
    - Produce a specific final description for the latents common in the examples, and what patterns you found.
    - Don't focus on giving examples of important tokens, if the examples are uninformative, you don't need to mention them.
    - Do not make lists of possible explanations. Keep your explanations short and concise.
    - The last line of your response must be the formatted explanation, using [EXPLANATION]:
\end{promptlisting}

\paragraph{Explanation scoring.}
\label{app:prompts:scoring}
The scorer assigns an integer match score from 0 to 4 for whether at least one highlighted token, in its surrounding context, matches the expert explanation.
We then binarise by treating scores ${\ge}\,2$ as positive.
A score of 2 is a plausible contextual match, so this counts incomplete matches as hits without treating mere token overlap (score 1) as a hit.

\begin{promptlisting}
You are a conservative classifier evaluating an explanation of when one Mixture-of-Experts expert routes a token.

The explanation can contain many unrelated activation modes connected by OR. In each tagged example, only token(s) inside <<double angle brackets>> are candidate routed positions. Judge whether at least one highlighted token, in its exact surrounding context, matches one specific mode in the explanation.

Assign an integer match score:
0 = no match; unrelated context.
1 = only superficial token, phrase, or broad-domain overlap.
2 = plausible but ambiguous or incomplete contextual match.
3 = clear, specific contextual match.
4 = direct or near-exact trigger-and-context match.

False positives are costly, do not give 3 or 4 merely because a match is possible, shares a topic, or uses a word named somewhere in the long explanation. Require the highlighted token's role and context to satisfy a concrete described pattern.

Return only one valid JSON object mapping every shown example id to an integer 0 through 4. Include each id exactly once and no other text.
\end{promptlisting}

\section{Example Expert Explanations}
\label{sec:appendix:expert_explanations}
\Cref{tab:expert_explanations} shows a RouterInterp explanation
for a randomly selected expert (Expert 2) from layer 12 of gpt-oss-20b.
The explainer is given activating windows grouped by SAE feature
and writes a single prose description of when the expert is selected.
We also show feature-level descriptions of a few of those SAE features
so the reader can inspect the micro-domains that make up the expert.

\begin{table}[t]
  \footnotesize
  \centering
  \begin{tabular}{lp{10cm}}
    \toprule
    \textbf{Level} & \textbf{Explanation} \\
    \midrule
    \textbf{Expert 2} & \small This expert activates across following distinct contexts: (1) Physical trauma and medical conditions—high activation on ``bruis'' and variants (``bruised'', ``bruises''), plus fragments like ``blister'', ``abras'', ``haemat'', ``infarct'', and ``stunned''. (2) Technical and structured writing—code, mathematical expressions, scientific notation, legal citations, and medical terminology from programming, mathematics, science, law, and academia, including punctuation and formatting elements. (3) Brand and term suffixes—final segments of brand names, company names, or technical terms before spaces or punctuation. (4) Electronic components—activation on ``diode'', ``resistor'', and ``rectifier'' in patents, diagrams, and technical specifications. \\
    \midrule
    \midrule
    \textit{Feature 41223} & \small \textit{Physical trauma terms—``bruis'' and variants (``bruised'', ``bruises''), plus fragments of injury-related terms (``blister'', ``abras'', ``haemat'', ``infarct'', ``stunned'') in clinical or descriptive contexts.} \\
    & \small \textbf{Activating Examples:} \\
    & \small [1] suffered scr\colorbox{green!20}{apes},\colorbox{green!100}{ bruis}\colorbox{green!37}{es}, bleeding \\
    & \small [2] . Her face was\colorbox{green!100}{ bruis}\colorbox{green!20}{ed} and there \\
    \midrule
    \textit{Feature 71261} & \small \textit{Programming and scientific notation—code snippets, structured data, and formal keywords with high activation on punctuation, symbols, and domain-specific terms.} \\
    & \small \textbf{Activating Examples:} \\
    & \small [1] \colorbox{green!73}{ public}\colorbox{green!49}{ Flight}\colorbox{green!20}{ execute}\colorbox{green!24}{()}\colorbox{green!20}{ throws}\colorbox{green!20}{ API} \\
    & \small [2] \colorbox{green!97}{77}\colorbox{green!20}{:}\colorbox{green!86}{58}\colorbox{green!100}{,}\colorbox{green!75}{78}\colorbox{green!47}{:} \\
    \midrule
    \textit{Feature 17613} & \small \textit{Structured context punctuation—punctuation, symbols, and function words in code, citations, mathematical expressions, and legal or technical documents.} \\
    & \small \textbf{Activating Examples:} \\
    & \small [1] g\textbackslash\{\}\colorbox{green!66}{|}\colorbox{green!36}{\_\{}\colorbox{green!100}{L}\colorbox{green!20}{\textasciicircum{}}\colorbox{green!73}{p}\} \\
    & \small [2] \colorbox{green!45}{,}\colorbox{green!53}{ Y}\colorbox{green!91}{.}\colorbox{green!75}{ S}\colorbox{green!100}{.}\colorbox{green!78}{ P}\colorbox{green!20}{iao}\colorbox{green!42}{ and} \\
    \midrule
    \textit{Feature 1738} & \small \textit{Brand and term suffixes—final segments of brand names, company names, or technical terms before spaces or punctuation.} \\
    & \small \textbf{Activating Examples:} \\
    & \small [1] Inverted microscope;\colorbox{green!100}{ Leica}\colorbox{green!20}{ Microsystems} GmbH \\
    & \small [2] viewed using a\colorbox{green!20}{ Le}\colorbox{green!100}{itz} Aristoplan microscope \\
    \midrule
    \textit{Feature 84536} & \small \textit{Electronic component names—``diode'', ``resistor'', ``rectifier'', and related abbreviations in circuits, patents, and technical specifications.} \\
    & \small \textbf{Activating Examples:} \\
    & \small [1] Q3, and\colorbox{green!100}{ diode}\colorbox{green!59}{ D}\colorbox{green!20}{1}. \\
    & \small [2] stations and as\colorbox{green!20}{ rect}\colorbox{green!100}{ifiers} for high-vol \\
    \bottomrule
  \end{tabular}
  \caption{
    RouterInterp produces an expert-level explanation from activating examples
    grouped by SAE feature.
    Feature-level descriptions below let the reader see how the expert's
    stated specialisations correspond to individual SAE features.
    For example, Expert 2 from gpt-oss-20b layer 12 specialises in physical trauma (\textit{Feature 41223}),
    programming and scientific notation (\textit{Feature 71261}),
    structured punctuation (\textit{Feature 17613}),
    brand-name suffixes (\textit{Feature 1738}),
    and electronic components (\textit{Feature 84536}).
    We display five features selected by gradient-based attribution.
  }
  \label{tab:expert_explanations}
\end{table}

\end{document}